\documentclass[acmsmall]{acmart}

\newcommand{\termsofservice}[1]{{%
ToS}}

\newcommand{\termsofserviceLong}[1]{{%
Terms of Service}}

\newcommand{\roberta}[1]{\texttt{RoBERTa}}
\newcommand{\privbert}[1]{\texttt{PrivBERT}}
\newcommand{\svm}[1]{\texttt{SVM}}
\newcommand{\rf}[1]{\texttt{RF}}

\newcommand{\REDACT}[1]{$\Box\Box\Box$} 

\newcommand{\redactCollege}[1]{[a U.S. University]}  

\newcommand{\briefCite}[1]{\citeauthor{#1} (\citeyear{#1})}

\newcounter{boldifyCounter}
\newcounter{fixmeSectionCounter}
\newcounter{fixmeTotalCounter}
\makeatletter
\@addtoreset{fixmeSectionCounter}{section}
\@addtoreset{fixmeSectionCounter}{subsection}
\@addtoreset{boldifyCounter}{section}
\@addtoreset{boldifyCounter}{subsection}
\makeatother

\newcommand{\boldify}[1]{}
\ifdefined\boldifyON
	\renewcommand{\boldify}[1]{
        \par\noindent
		\stepcounter{boldifyCounter}
		\textbf{{\color{green}**}
		~\arabic{section}.\arabic{subsection}.\arabic{boldifyCounter}
		: #1} 
	}
\fi 

\newcommand{\reportOnFIXME}{%
    \newcount\iterCounter
    \iterCounter=1
    \newcount\endCounter
    \endCounter=\totvalue{fixmeTotalCounter}
    \advance \endCounter +1
    There are 
    {\color{red}\total{fixmeTotalCounter}} 
    FIX\_ME\\
    links:
    \loop
        \hyperlink{fixTag\the\iterCounter}{\#\the\iterCounter}
        \advance \iterCounter +1
    \ifnum \iterCounter < \endCounter
    \repeat
}

\newcommand{\FIXME}[1]{} 
\ifdefined\fixmeON
	\renewcommand{\FIXME}[1]{\par\noindent
		\stepcounter{fixmeSectionCounter}\stepcounter{fixmeTotalCounter}
		{\color{red}\fbox{\color{black}
			\parbox{.965\linewidth}{
				\textbf{\hypertarget{fixTag\thefixmeTotalCounter}{FIXME}	\arabic{section}.\arabic{subsection}.
        		\arabic{fixmeSectionCounter} (\color{red}
        		\#\arabic{fixmeTotalCounter}):} #1}}
        }
	}
\fi 

\newcommand{\FIXED}[1]{}
\ifdefined\fixmeON
	\renewcommand{\FIXED}[1]{\par\noindent%
		{\color{black}\fbox{\color{black}%
			\parbox{.99\columnwidth}{%
				\color{blue}#1}}%
        }
	}
\fi 

\newcommand{\draftStatus}[1]{}
\ifdefined\draftStatusON
	\renewcommand{\draftStatus}[1]{
        \hfill **#1
	}
\fi 

\usepackage{soul}
\usepackage{calc, enumitem} 
\usepackage{hyperref}
\usepackage{booktabs}
\usepackage{array}
\usepackage{placeins}
\usepackage{totcount}\regtotcounter{fixmeTotalCounter}
\usepackage{datetime2} 

\setcopyright{acmcopyright}
\copyrightyear{2024}
\acmYear{2024}
\acmDOI{99.9999/9999999.9999999}
\acmJournal{TOPS}
\acmVolume{99}
\acmNumber{99}
\acmArticle{99}
\acmMonth{99}

\begin{document}
\title[An Automated Approach for Summarizing and Analyzing Overlaps in Privacy Policies and Terms of Service]
{Demystifying Legalese: An Automated Approach for Summarizing and Analyzing Overlaps in Privacy Policies and Terms of Service}

\author{Shikha Soneji}
\author{Mitchell Hoesing}
\author{Sujay Koujalgi}
\author{Jonathan Dodge}
\email{[sxs7000, mdh5934, snk5290, jxd6067]@psu.edu}
\affiliation{%
\institution{The Pennsylvania State University}
\streetaddress{Westgate Building}
\city{University Park}
\state{Pennsylvania}
\postcode{16802}
\country{USA}}

\renewcommand{\shortauthors}{S.\ Soneji, M.\ Hoesing, S.\ Koujalgi, and J.\ Dodge}

\newcommand{\abstractTextOLD}{%
The convoluted legalese language coupled with ambiguous use of terminology makes it difficult for people to comprehend terms and policies for products and services that they use.
Thus, they become legally bound to a contract by agreeing to and accepting the terms presented to them but are unaware of the contract's contents, which can lead them to give information without realizing how the service provider will use it.
Current state-of-the-art summarizing and scoring tools for policy documents such as Terms of Service and Privacy Policy documents rely heavily on manual annotations that break down the legalese into layman's terms.
Our ultimate goal is to provide simplified information and scores for policy documents on-demand in an automated fashion, allowing users to better understand contracts for new services that are not yet annotated.
To that end, this paper describes a variety of language models to automate this analysis and the dataset we collected to train and fine-tune our models.
Our tool reduces the verbosity and complexity of the legalese present in policy documents in an attempt to simplify the contents.
This study also aims to identify overlaps of content in the different documents that the GDPR requires to be mutually exclusive of each other to avoid redundancies.
While many of the privacy policies and terms of services documents we studied were free from overlap, some contained the same contents, indicating violations of GDPR guidelines. Among all the transformer-based and traditional models used in the paper, Roberta performs the best with a superior F1-score (0.74). 
}

\begin{abstract}
The complexities of legalese in terms and policy documents can bind individuals to contracts they do not fully comprehend, potentially leading to uninformed data sharing.
Our work seeks to alleviate this issue by developing language models that provide automated, accessible summaries and scores for such documents, aiming to enhance user understanding and facilitate informed decisions.
We compared transformer-based and conventional models during training on our dataset, and RoBERTa performed better overall with a remarkable 0.74 F1-score. 
Leveraging our best-performing model, RoBERTa, we highlighted redundancies and potential guideline violations by identifying overlaps in GDPR-required documents, underscoring the necessity for stricter GDPR compliance.
\end{abstract}



\maketitle

\section{Introduction
\draftStatus{1.5}}
\label{sectionIntro}

\boldify{\termsofservice{}'s are commonplace and accepting them is a part of daily life, BUT people neither read nor understand them, which is a problem}

Users typically agree to lengthy and complicated \termsofserviceLong{} (\termsofservice{}) contracts without fully understanding them, making it difficult to keep track of changes to policy~\cite{obar2020biggest, auxier2019americans}.
This calls into question the reasonableness of expecting users to make informed decisions given that they do not comprehend the terms.

\boldify{Simplifying the document both in length and complexity may prove beneficial to users.
Imagine a pipeline (fig1) that works like this: @describe a simplification/scoring pipeline@.
}

A technology that can automate the simplification and categorization of popular \termsofservice{} documents would be immensely beneficial, enhancing user understanding of accepted policies and facilitating the identification of concerning changes.
We envision an automated system that begins with the text of a \termsofservice{} document for a new product or service.
The prospective user copies and pastes the text into an automated tool, which extracts key concepts and then presents some information in a format that is shorter and easier to read, such as a numeric/letter score alongside a bullet list of the most important concepts.

\boldify{At a high level, we work toward that goal in the following way.
First, we narrowed our focus to website documents and assembled a data corpus: @describe the document corpus and its human annotation origins@}

Our work focuses on extracting key concepts from a data corpus we scraped from \citeauthor{tosdr} (ToS;DR)~\cite{tosdr}.
This tool translates digital service regulations based on  annotations by community members, which admins then authorize or reject.
It includes service names, condensed policy details, implications, and a user-friendliness score.

\boldify{With our data corpus, we then approached our first task: @describe TASK 1 - case analysis@}

For the automated annotation of policy documents, we have performed multi-class text classification on sentences, with a label space of size 246.
The output classes each represent a key concept from the ToSDR taxonomy, such as \emph{``The service has a no refund policy''}.
We refer to each of these concepts as a \textbf{case}, and so this defines our first task \emph{Case Analysis}.
This task extracts key concepts from policy documents, in a manner ready for scoring.

\boldify{With the same document corpus, we can also make progress on a second interesting task: @describe TASK 2 - docType analysis@}

Since our data corpus includes different \textbf{document type}s, we are also able to define a second task \emph{Document Analysis}.
Again, we perform multi-class text classification on sentences, but this time with 5 labels.
Here, the output classes each represent a type of policy document (i.e., docTypes are: \termsofserviceLong{}, Privacy Policy, Cookie Policy, Data Policy, and Other Policy).
The reason we are interested in this task is that GDPR guidelines~\cite{gdprGuidelines} describe disjoint sets of concepts that should appear in privacy policy and \termsofservice{} documents, to avoid redundancies.
One way to determine if policy document writers are following the guidelines is to measure how well classifiers can predict the document type of the source (akin to Pozen et al.'s work~\cite{pozen2019} predicting political party affiliation from text).

\boldify{To attack these two problems, we use the following models: @describe the MODELS, 4x@}

In both tasks, we utilized four machine learning models: RoBERTa~\cite{liu2019roberta}, a transformer-based model optimized for performance; PrivBERT~\cite{srinath2021privaseer}, a BERT variant tailored for privacy-policy tasks; a Linear Support Vector Machine (SVM)~\cite{cortes1995support}, renowned for efficient high-dimensional data handling; and the Random Forest~\cite{breiman2001random}, an ensemble learning method praised for its versatility.
Together, these models provided a comparison between state-of-the-art transformer models and more conventional machine-learning approaches.
Note that our study does not include any LLMs for explainability reasons, discussed further in Section~\ref{secDiscussion}.

\boldify{Using the best-performing model from the experiments presented, we then sought to understand the extent to which websites comply with GDPR guidance by doing the following: @describe TASK 3 - overlap analysis@}

Next, we select the best-performing model for evaluating concept overlap across document types using two main lenses.
First, we evaluate the document type classifier's ability to distinguish the sources of phrases derived from distinct document types~\cite{pozen2019}.
It shows that machine-learning classifiers can predict a speaker's political membership based just on the semantic content of a constitutional utterance in Congress through their work on measuring polarization.
We use a similar approach, measuring pairwise accuracy in our classifications to quantify the conceptual overlap across different categories of documents.
For example, due to the smaller conceptual overlap between these clearly distinct document types, the classifiers would probably perform better if we compared privacy rules to cookbooks rather than terms of service.
Second, we utilize a case classifier to identify instances this classifier outputs originating from distinct document types, providing another measure of concept overlap.
This multi-pronged approach allows us to triangulate  results from both overlap measures.

Our principal contributions are listed below:
\begin{enumerate}
    \item Application of Concept Overlap Measurement in a New Domain: Our goal is to objectively quantify and emphasize overlap in policy documents by extending methods from ~\cite{pozen2019} to legal texts.
    This can help regulators, customers, and authors.
    \item  First Moves in the Direction of an Automated Analysis Tool: The existing policy document analysis methods are restricted to known data and frequently depend on human annotations.
    Our method opens the door for a tool that can examine fresh, unexplored data.
    \item Empirical Analysis of NLP Models: We examine transformer-based and conventional models to evaluate their suitability for developing a practical policy document analysis tool, emphasizing the promise of tiny language models.
\end{enumerate}

This paper explores the following RQs:
\begin{enumerate}[leftmargin=30pt, nosep]
    \item[\textbf{RQ1}] \emph{Case Classification} - What is our technique's efficacy in classifying cases based on a human-curated taxonomy for facilitating policy document simplification?

    \item [\textbf{RQ2}] \emph{Overlap Quantification} - How much concept overlap exists among types of policy documents?

    \begin{enumerate}[leftmargin=5pt, nosep]
    
        \item[\textbf{2a}] How well can the docType classifiers differentiate the source of sentences originating from different policy documents?
    
        \item [\textbf{2b}] To what extent do the Case Classifier's outputs for \termsofservice{} and Privacy Policies differ?
    
    \end{enumerate}

    \item [\textbf{RQ3}] \emph{Cases that Overlap} - What can we learn by inspecting the output of the Case classifier for different docTypes?
    
    \begin{enumerate}[leftmargin=5pt, nosep]

    \item [\textbf{3a}] Which cases exhibit the greatest (or least) overlap?

    \item [\textbf{3b}] Which type of document is encroaching and which is subject to encroachment?
    
    \end{enumerate}
\end{enumerate}

\section{Background and Related Work
\draftStatus{D1.5}}

\subsection{People's Understanding of Legal Documents
\draftStatus{D1.5}}

\boldify{People don't read these documents, here is evidence}

The \emph{``biggest lie on the internet''}~\cite{obar2020biggest} is, of course, \emph{``Yes, I have read and agree to the terms.''}.
Even though there are no rules requiring websites to disclose their \termsofservice{}, a number of laws may call for these declarations.
Therefore, before using the services provided, users must read and comprehend the \termsofservice{} provided by an organization.
While it should have taken them 15–17 minutes to read a \termsofservice{} document thoroughly, users only spent an average of 51 seconds doing so~\cite{obar2020biggest}, indicating that information overload was a substantial factor influencing their reading behavior.
Companies that use extremely restrictive \termsofservice{} may struggle to sustain their consumer base and overall profitability~\cite{bakos2014does}

\boldify{Even those who read it don't understand it, here is evidence}

Despite some users' claim to have read privacy rules, only 22\% of users say they have done so thoroughly, with the majority admitting to having merely skimmed or read a section~\cite{auxier2019americans}.
If consumers lack comprehensive understanding and providers know their policies largely go unread, the question arises as to how users can make informed decisions.

\boldify{Here is a bit about WHY they don't read it Reason 1: some are fine with the tradeoff}

To ascertain if consumers read \termsofservice{}, the researchers in ~\cite{maronick2014consumers} performed an empirical analysis, discovering that levels of concern about how data is used by services or websites they often use might range widely among people.
Some individuals may feel highly or somewhat at ease with companies utilizing their personal data to enhance their fraud prevention measures~\cite{auxier2019americans}.
According to the research in~\cite{lippi2019claudette}, 50\% of Americans feel at least somewhat comfortable with corporations exploiting their personal information to develop new goods, while 49\% are highly uncomfortable.

\boldify{Reason 2: people are scared off by the obfuscation.
NLP can help cut through the noise, but unclear to what avail.}

The length of policy documents, as well as the legal jargon used, give the appearance that the documents are not intended for readability by laypersons~\cite{robinson2020beyond}.
To borrow a phrase, it seems that the use of legalese \emph{``obfuscates or impresses rather than clarifies''}~\cite{lipton2018troubling}.
To address such concerns, various researchers have sought to condense the material to a level that is easy to read without sacrificing detail, with the help of Machine Learning techniques such as summarization~\cite{tesfay2018privacyguide}, automated annotation~\cite{adhikari2020automated} and classification~\cite{zimmeck2014privee}.
However, it is not clear that users will find machine-generated analysis trustworthy.

\boldify{trustworthy you say, sounds like XAI}

\boldify{...and failure can be dire, here are some horror stories}

Numerous studies have discovered damage to consumers' trust in a company as a result of websites' failure to protect personal information~\cite{pilton2021evaluating}.
Some websites have a history of abusing users' trust by sharing, selling, or researching personal information.
More than 51\% of participants surveyed said that having Google or Yahoo track their search activity affected their mood ~\cite{melicher2016preferences}.
Whether or not users comprehend the \termsofservice{}, it has significant legal ramifications.
A federal appeals court determined that a client was responsible for credit card fees that he was not aware had accrued because they were stated on an online ``Terms and Conditions'' page that he did not read, despite checking a box indicating that he had done so~\cite{2012davis}.
Even with plain-language translations of legalese, the efficacy of such approaches may be undermined by a lack of user motivation to understand the \termsofservice{}.

\subsection{Mining and Annotation of Legal Documents
\draftStatus{D1.5}}

\boldify{So far we have mostly focused on end users looking at a policy document and what data scientists might do to help that user.
Now let's switch focus to the data sources and describe a few of the key concepts in data mining that underpin this work, due to processing a document corpus.}

So far we have focused on end users looking at a policy document and what data scientists might do to help that user, now we switch focus to the data sources themselves.
An early tool that finds service exceptions at the phrase level from a vast corpus of legal contracts is called Contract Miner~\cite{gao2011mining}, using a simple but effective unsupervised information extraction method.
Contract Miner begins with pre-processing, followed by applying linguistic patterns and parsing service exception phrases.
Privee~\cite{zimmeck2014privee} builds on ToSDR by combining crowd-sourcing with rules and machine-learning classifiers to identify privacy regulations that do not yet have reviews in the crowd-sourcing repository.
In their study, the authors introduced ``PrivacyCheck''~\cite{zaeem2018privacycheck}, a Chrome browser extension leveraging data mining models to automatically generate concise summaries of online privacy policies, demonstrating its effectiveness across 400 companies' policies with an accuracy rate of 40\% to 73\%, and currently adopted by over 400 Chrome users.
This study ~\cite{bui2021automated} investigates the use of automated methods to extract and present data management practices from privacy policies. Their research focuses on the effective parsing of complicated legal texts using natural language processing and machine learning approaches, with the goal of increasing transparency and user understanding of privacy practices.

The work in ~\cite{amos2021privacy} compiles a large dataset of over a million privacy regulations and examines their evolution over time. This unprecedented scale reveals insights into how data security procedures have evolved, highlighting trends, regulatory compliance, and the influence on consumer privacy awareness. Their work distinguishes out due of its broad temporal scope.
\boldify{As we get closer to related work, what is state of the art in mining, annotating, and processing "LEGAL DOCS IN GENERAL" (i.e., not restricted to \termsofservice{} and such)}
Researchers present a technique to automatically scan and annotate privacy regulations using natural language processing in their study~\cite{alabduljabbar2022measuring} on evaluating privacy threats of websites offering free content.
This strategy exposes important information about online privacy practices and demonstrates how automated technologies may effectively support transparency in the field of digital privacy.
Kost and Freytag~\cite{kost2012privacy} take a different approach to examining privacy practices for compliance using ontological models.
By offering a thorough tool for automatic privacy evaluation, this approach improves knowledge of data management and protection and is essential for navigating the complexity of online privacy.
Extending this work, Liu et al.~\cite{liu2021have} presents a way to assess privacy policies' GDPR compliance through natural language processing. PolicyQA, a dataset designed for reading comprehension exercises on privacy policies, is introduced by \cite{ahmad2020policyqa}. By offering a systematic framework for querying and interpreting privacy regulations, improving machine understanding, and enabling more easily available privacy information for users, this resource seeks to support breakthroughs in natural language processing applications.
It draws attention to discrepancies in existing procedures and provides a scalable way to boost corporate compliance, accountability, and transparency in data processing.
The work of ~\cite{ahmad2021intent} focuses on the use of natural language processing (NLP) techniques to slot filling and intent categorization in privacy regulations. By using an innovative technique, they want to improve automated privacy policy analysis and comprehension by better understanding and classifying the purposes behind data handling activities and identifying specific information slots within the rules.
\boldify{And now we arrive at the lowest level of specificity, policy documents, and where we get our labels (i.e., human annotations, so background/data source info, NOT related work, which is where we see people automating stuff).}

The project (ToSDR)~\cite{tosdr} aims to address the problem of customers skipping through \termsofservice{} by having volunteers from the online community review, discuss, and evaluate various policy papers.
Assessments are available online and through free browser add-ons.
This crowdsourcing platform allows anyone to participate by annotating any policy-related document, which the ToSDR team subsequently evaluates.
As such, while it provides substantial coverage, there might be occasional gaps concerning the latest policies or those from less popular services.
The OPP-115 dataset~\cite{wilson2016creation}, encompassing 115 privacy policies from US companies annotated by law students, has facilitated privacy studies and tools.

\subsection{Related Work
\draftStatus{D0.8}}

\boldify{Nearest neighbors on NLP pipelining are those we lay out in this section.
Broadly speaking, we  will follow the history of improvements shown in a recent historical analysis}

The current state of privacy policy analysis tools shows a range of advantages and disadvantages.
Research conducted by Wagner et al.~\cite{wagner2023ages} presents a comprehensive analysis of 25 years of privacy policies using machine learning, highlighting post-GDPR improvements, such as aiming to make policies more user-friendly and aligned with privacy preferences.
This section collates past and current efforts at implementing various tools and services to make the contents of the policy documents more accessible to the general public and user-friendly to educate them about the contents of the policy documents.

Utilizing natural language processing (NLP), machine learning, and crowdsourcing, the Usable Privacy Project~\cite{sadeh2013usable} successfully combines NLP, machine learning semi-automatically annotates privacy laws.
Its objective is to simplify and improve the usability of privacy settings, tackling the problem of lengthy and complex restrictions that frequently impede people's ability to comprehend and take control of their data.

\boldify{There are some tools out there which automate classification or summarization, but they stick to things they have "tested", so to speak; such tools are:...}

Polisis~\cite{harkous2018polisis} and PrivacyCheck~\cite{zaeem2021large}, two of the most recent publicly available programs, utilize machine learning for privacy policy summarization.
Polisis extends its utility by offering educational materials and visualizations on privacy.
However, these tools mainly generate summaries and do not offer structural analysis of policies or examine relationships between different documents and cases.
Furthermore, their analysis is usually limited to a certain number of websites, creating a `walled garden' effect that lacks broad scalability.
Although constructed using a sophisticated combination of machine learning and NLP approaches and demonstrating a noteworthy accuracy of 0.74 in classifying privacy policy information, PrivacyGuide~\cite{tesfay2018privacyguide} is presently dealing with maintenance and potential activation difficulties.
Despite being an originally optimistic technique inspired by the EU's GDPR, this restricts its utility.
This project also exercises the boundaries of the documents uploaded and tested so far and does not apply to `unseen' documents.

\boldify{Our tool has an educational element, so here's what we know about other mechanisms for educating people about documents.}

The work of Ravichander et al.~\cite{ravichander2019question} improves question-answering systems for privacy rules by bridging the legal and computational domains.
Their research presents an innovative method to analyzing privacy regulations by fusing legal expertise with natural language processing tools, with the goal of improving public understanding and accessibility to these rules.
PrivExtractor~\cite{bolton2023imbalance} is a dashboard designed to help users better understand virtual assistant data privacy policies which shows that although businesses follow the law, complete openness in data use is not guaranteed by rules.
Along the same educational lines, Golbeck et al.~\cite{golbeck2016} studied user awareness of the personal data shared with Facebook apps and found that consumers were initially unaware of the degree of data access these apps might obtain.
Those authors compared effectiveness of different methods of learning about the information Facebook apps can access (reading the document, watching a ``horror''-themed video, an app, and all of the above), and generally found that the video was among the most effective strategies.

\boldify{And now we reach the most similar thing to ours, which is two things: Lukose and Barrister}

The two works most similar to ours are as follows.
Perera and Perera~\cite{perera2021barrister} have recently proposed a framework called `Barrister' to simplify online legal documents, focusing on class action waivers, yet it risks oversimplification and neglect of other critical policy aspects.
By leveraging advanced processing techniques, Barrister aims to distill essential information and present it in a concise, user-friendly format, thereby enhancing transparency and aiding users in making more informed decisions regarding their privacy and the use of digital services.
Simultaneously, classification and summarization of \termsofservice{} highlights privacy pitfalls~\cite{lukose2022privacy}, offering a novel model that combines extractive-classifier-abstractive approaches in a hybrid manner to address the complexity of online terms of service.
By curating a dataset via online scraping, this model benefits the community and outperforms existing approaches despite a considerably smaller training dataset.
It also effectively highlights and contextualizes important privacy and data collection/use sections within these documents.

\boldify{Now we describe how our thing is cooler}

By taking a sentence-level ``concept classification'' (e.g., case or docType) approach as laid out in Section~\ref{sectionIntro}'s RQs and Section~\ref{sectionMethod}'s Methodology, we seek to \textit{simplify} documents without relying on \textit{summarization} to be more robust for unseen data.
One consequence of building concept classifiers is that it is very natural to apply them to compare/contrast concepts present in policy documents: highlighting overlapping concepts, identifying encroachment, and analyzing case-wise overlaps.
The proposed research framework holds potential to add significant value to the existing collection of policy analysis tools.
The three primary merits of our tool are its accuracy, scalability, and generalization capabilities.
Unlike existing tools, which are currently limited to analyzing around 400 websites, our tool would be scalable and can analyze \termsofservice{} for nearly all websites in various domains.
This broad scalability means our tool can offer a more accurate, comprehensive, and inclusive analysis enhancing the validity and reliability of our tool's analysis and providing a robust foundation for policy interpretation and enforcement.

\section{Methodology
\draftStatus{D0.9}}
\label{sectionMethod}

\subsection{Stage 1: Data Collection
\draftStatus{D1.0}}

\boldify{Stage 1 - Let's go get data! We scraped @kind of data@ from @sources@ using @scraper@.
First, let's define input variables and output variables for each of the three tasks.}

In a two-step data collection process, we first used the ToSDR~\cite{tosdr} streaming API to systematically crawl and scrape data in the third quarter of 2023 to make sure that our dataset remained up-to-date.
Next, we extracted information from the ToSDR website via BeautifulSoup~\cite{beautifulSoup}, resulting in the scraped data saved in human-readable .CSV format.
This dataset contains a wealth of attributes, including \textbf{Description}, \textbf{Case}, Title, Status, Comments, and Author ID.
Our system uses \emph{Description} as input to both the Case and DocType Classification tasks in the training exercises, while during testing, the models output a \emph{Case} or \emph{DocType} for each \emph{Description}, respectively.

To elaborate on the dataset specifics, \emph{Description} refers to the text drawn from the service's policy document.
The \emph{Case}, on the other hand, is a categorical label applied to a Description that is previously defined by the ToSDR team.
This team has impressively formulated about 246 such cases, which can serve as a ``summary'' of the key concepts in the document.
The \emph{Title} contains a categorical label signifying the type of policy document in which the Description appeared, such as \termsofserviceLong{}, Privacy Policy, Cookie Policy, Data Policy, and any Other Policy.
The \emph{Status} contains a categorical variable Pending/Accepted/Declined/Changes Requested.
The webpage also includes the \emph{Author ID} of the contributor as well as \emph{Comments} from the ToSDR team or the author.

\subsection{Stage 2: Data Preprocessing
\draftStatus{D1.0}}

\begin{figure}[t]
    \centering
    \includegraphics[width=\linewidth]{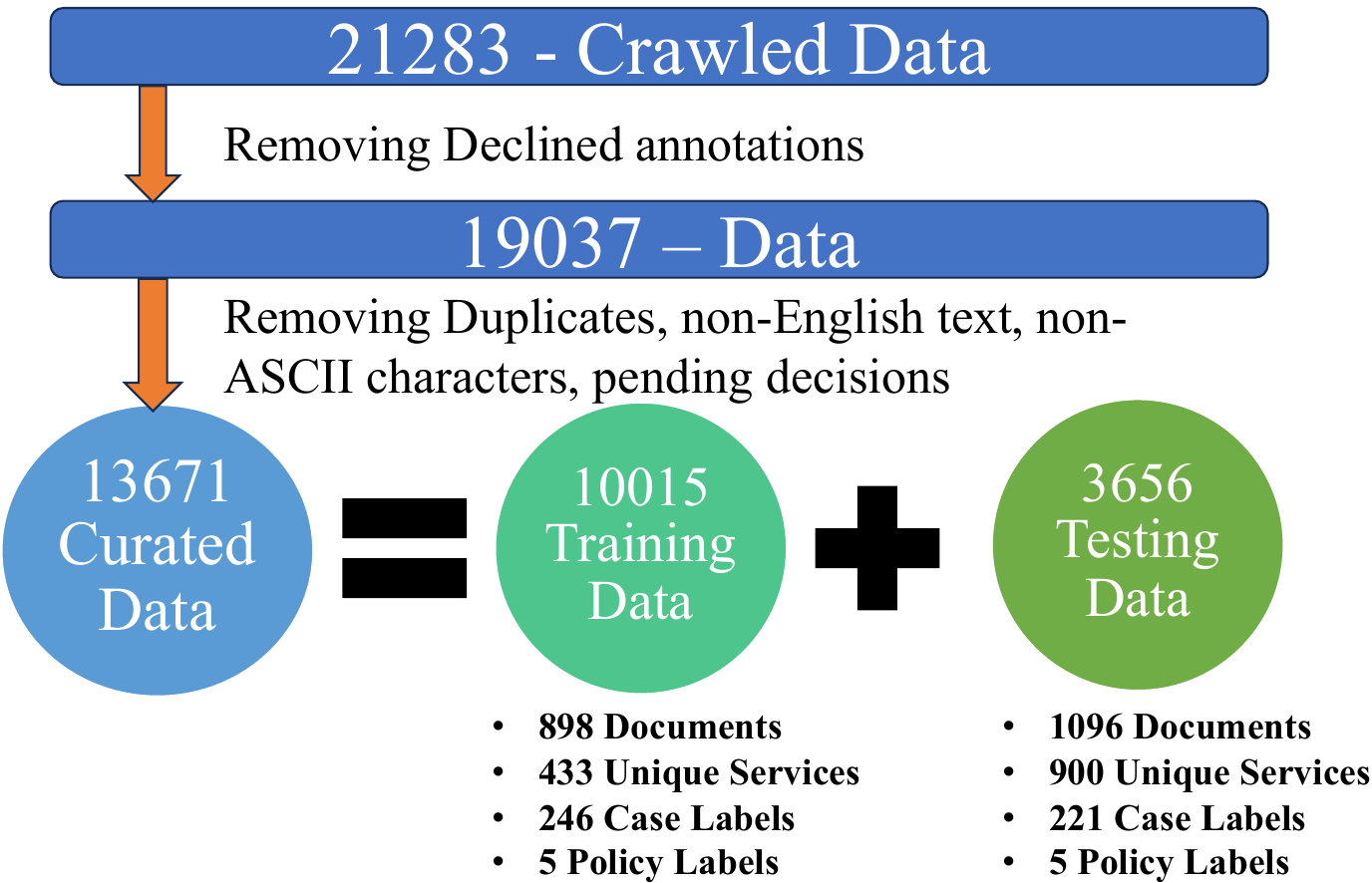}
    \caption{Data Dissection}
    \label{fig:dissection}
    \vspace{-10pt}
\end{figure}

\boldify{Stage 2 - Prepare data for training, as per Figure~\ref{fig:dissection}.
After the initial scrape, we had @XYZ@ documents from @MNO@ services, for a total of @ABC@ sentences.
We also did the following to prepare for training (removed HTML tags, removed non-ASCII characters(?), mapped self-described docTypes to our narrower group of docTypes, removed XYZ to prevent data leakage, removed duplicates, and whatever else}

Our data dissection is depicted in Figure~\ref{fig:dissection}.
Using the ToSDR streaming API, we collected 21,283 total annotations.
The trimming resulted in a curated dataset of 13,671 items.
We mapped the document types down to 5 labels for consistency, as numerous websites may have their own titles for privacy documents.

\boldify{Next up is a test/train split}

We then partitioned the refined samples into two datasets.
The training dataset consists of services that boast at least 10 annotations per document, while the test dataset has fewer than 10 annotations per document.
Our dataset's test-train split of 73–27 is based on the sparsity of label application, where annotations are frequently missing where they should be.
Based on our hypothesis, documents with more annotations ($>10$ annotations) probably have fewer missing labels, which is why we included them in our training set.
After thresholding, 896 training documents and 1096 testing documents are produced.

\begin{figure*}[htb]
    \centering
    \includegraphics[width=.5\linewidth]{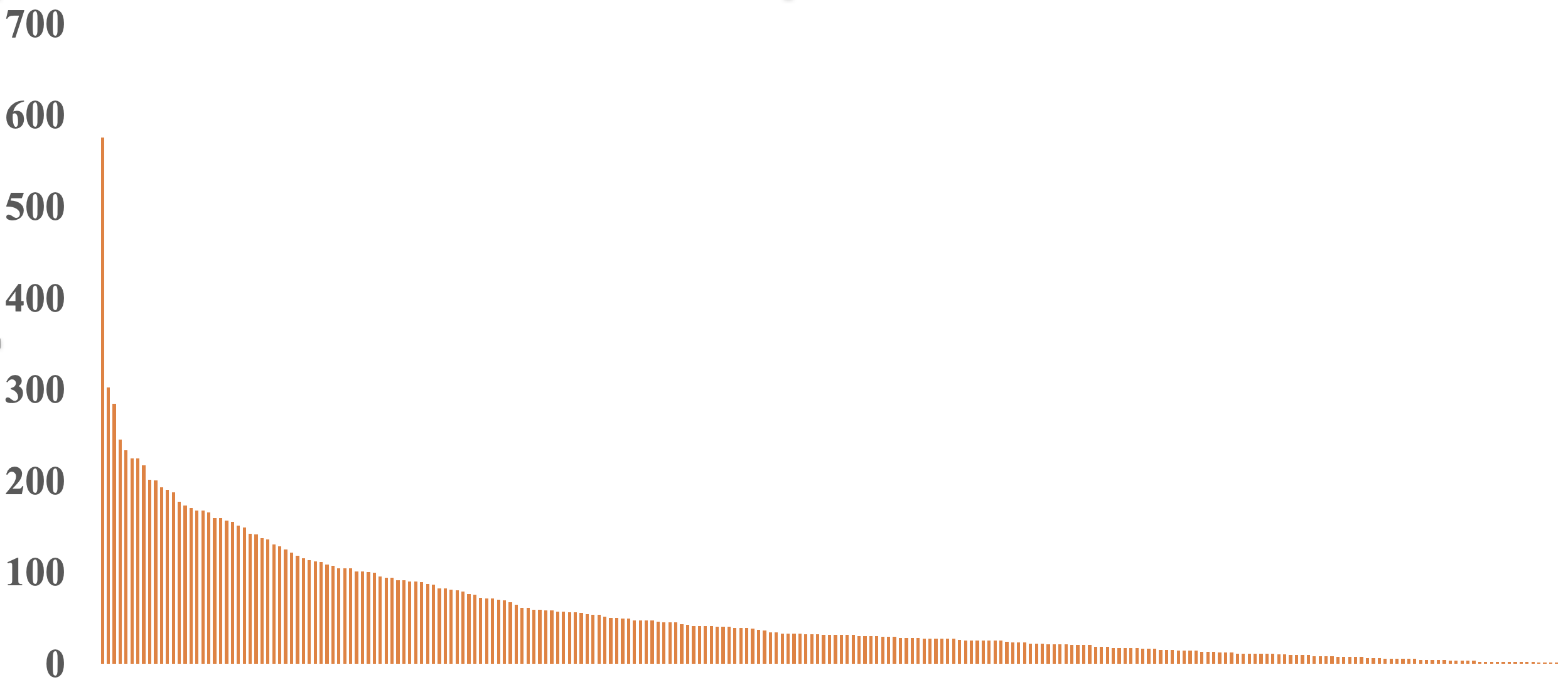}
    \hfill
    \includegraphics[width=.4\linewidth]{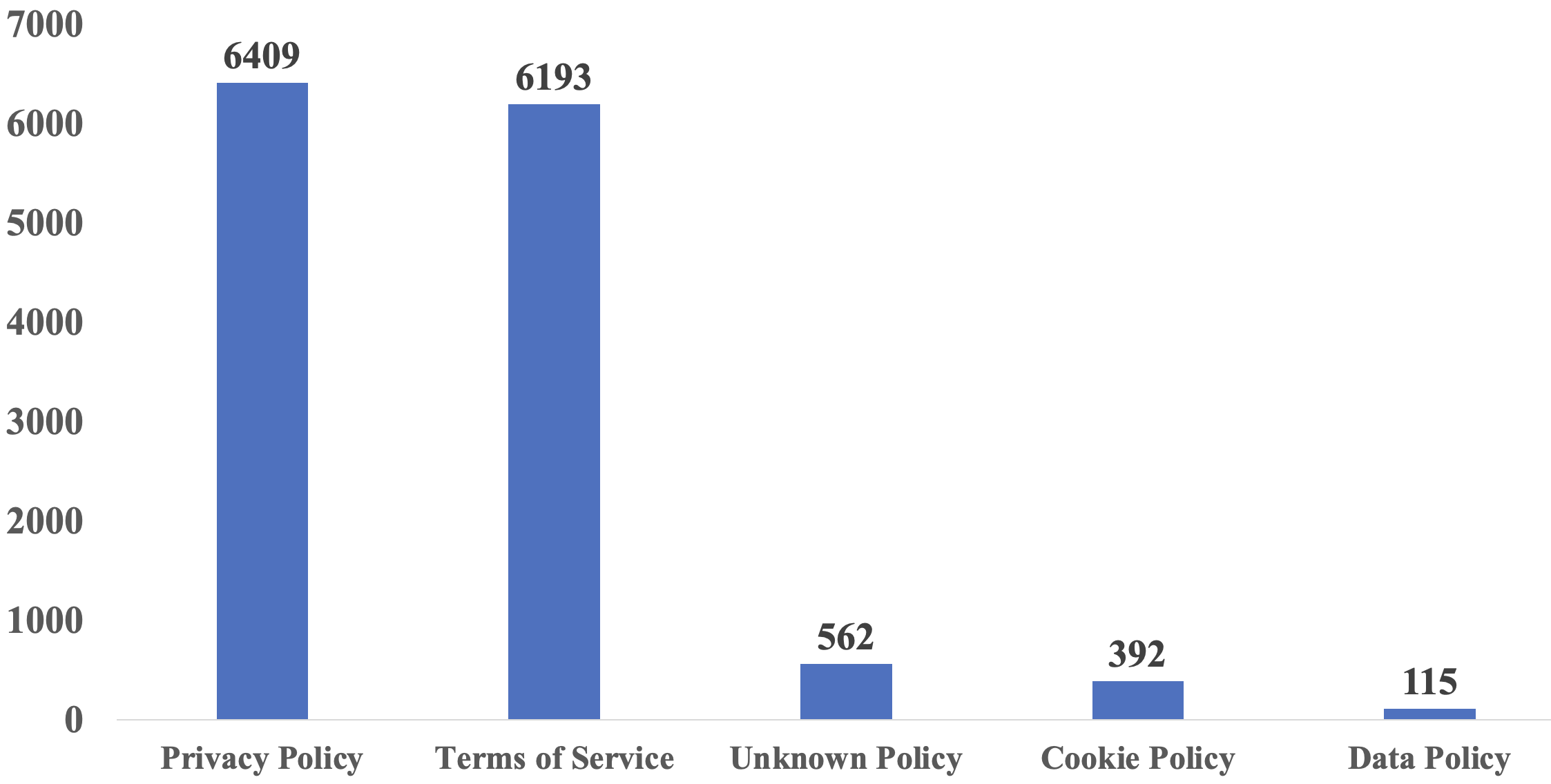}
    \caption{Distribution of examples over our labels for...
    (\textbf{left:}) Cases, consisting of 246 labels with frequencies varying 575--1; and
    (\textbf{right:}) DocTypes, consisting of 5 labels with frequencies varying 6409--115.}
    \label{fig:longTail}
\end{figure*}

\boldify{This dataset exhibits some rather severe issues with long-tail distributions, as visible in Figure \ref{fig:longTail}.
To assist with problems resulting from this distribution we mostly set up sampling strategies in the preprocessing step}

The ``\termsofservice{}'' and ``Privacy Policy'' classes account for 92\% of the policy multi-class data in the ToSDR data, as seen in Figure~\ref{fig:longTail}.
By using oversampling with replacement to balance the distribution of each class, we were able to alleviate the imbalance that existed.
We applied oversampling to the train set only.
To take a concrete example with the docType classification task, ``Cookie Policy,'' ``Data Policy,'' and ``Unknown Policy'' had 291, 99, and 363 samples, respectively, compared to 4,728 for \termsofservice{}, and 4,534 for Privacy Policy.

The oversampling approach we took was rather naive, so it is not a main focus of the paper, rather an attempt to mitigate a problem that is obvious upon visual inspection of our label distributions.
After oversampling, the total sample count for the docType data grew from 10,015 to 23,640, guaranteeing that each of the five classes had 4,728 occurrences.
The work on case classification used a similar method.

One concern we had using oversampling was the increased chance of overfitting the model to the oversampled data classes.
To evaluate this concern we consider our Confusion Matrix, which appears later in Section~\ref{secOverlap} as Table~\ref{tab:confusion}.
We observed, in example, that even though we heavily oversampled ``Data Policy'', the accuracy decreased.
The same is true for the ``Other Policy'' and ``Cookie Policy'' classes.
These results alleviated our concerns of overfitting.

\subsection{Stage 3: Train Machine Learning Models
\draftStatus{D1.5}}

\boldify{We chose to evaluate the following 4 models: @list them@. The reasons we went with these ones and not others are @whatever reasons@. The first 2 are transformers}

In our multi-class classification tasks, we employ four notable models: PrivBERT~\cite{srinath2021privaseer}, RoBERTa~\cite{liu2019roberta}, Linear SVM, and Random Forest (RF).
With PrivBERT's specific tailoring for privacy rules and RoBERTa's dynamic token masking and larger training dataset, both models' transformer-based architectures effectively manage long dependencies and complex structures common to intricate policy documents.

\boldify{The second 2 are not transformers, and we mostly chose them for the case classification task. We use the same ones for the doctype classification task}

The high number of classes (246) of the Case Classification task is handled by our third model, the Linear SVM classifier, which also offers resistance to overfitting. The RF model uses the same models for the docType Classification task for simplified presentation while simultaneously managing a vast feature space and possible correlations with built-in overfitting avoidance. 
We chose to utilize the same model version/configurations when possible to keep the tasks  commensurate to simplify comparison

\boldify{In an effort to be fair to each model, we set hyperparameters as follows: oversample/none was a variable we manipulated. }

In essence, `Normal' and `Oversample' represent two different strategies for handling class imbalance in the dataset, each with its unique implications on the model's training and performance.
In the context of this experiment, `Normal' refers to the original, unmodified set of training and testing data.
This is the base dataset that maintains its initial distribution of classes, thereby providing a direct reflection of the real-world scenario, including any class imbalance inherent to the phenomenon and its detection.
In contrast, `Oversample',  signifies our attempt to address the issue of an imbalanced dataset, specifically within the realm of long-tailed distribution problems.
In such cases, certain classes possess significantly more samples than others, potentially causing the model to bias towards the majority class.
We trained our NLP models (RoBERTa and PrivBERT) for 5 epochs with warm-up steps for the learning rate set to 500 and weight decay set to 0.01.
Based on our computing resources, we set the batch size for training to 16 and for the test to 64. 

\subsection{Stage 4: Evaluate Models and Measure Overlap
\draftStatus{D1.5}}
\label{sec:overlap}

\boldify{Having trained a variety of models, we wanted to select the best one to perform TASK 3.
To do so, we examine F1 scores because @make reasons explicit@.
We focused on pairwise overlap in these two docTypes because they have the most examples and also GDPR source says they should be different in the following ways.
}

To perform the overlap measurement, we need to select the best classifiers from those available.
It is apparent that F1-score serves as a more comprehensive metric than accuracy because accuracy can be misleading while working with imbalanced datasets.
We have used weighted F1 scores for all the metrics on both tasks.
Table~\ref{tab:confusion} contains scores for the individual docTypes. We use macro-averaging to construct evaluation metrics because it offers equal weight to each class while doing multi-class classification.
It is important to clarify that the F1 score does not measure conceptual overlap between document types.
Instead, our approach utilizes pairwise accuracy for this specific purpose.
This method reflects the likelihood of correctly identifying the document type based on semantic content, similar to how political affiliation prediction can measure conceptual overlap in speeches~\cite{pozen2019}.
Pairwise accuracy effectively captures the nuances in different document types, thereby indicating the degree of overlap more accurately than the Macro-averaged F1 score.

\boldify{We empirically determined our best available model for case prediction, then went about measuring overlap via that model.
To measure the overlap, we used both the case and doctype classifiers in the following way...}

Once we determined the best classifiers for predicting cases and docTypes, we set about to measure concept overlap between docTypes in two ways.
First, we use the best performance attained on docType classification as a proxy for measuring the conceptual overlap between docTypes, following methodology in prior work~\cite{pozen2019}.
Second, we run the case classifier and examine the frequency at which it detects a particular case in each docType.
If there is no concept overlap, then we should anticipate a high performance on docType classification and a few cases with large numbers of occurrences in each docType.

\boldify{Having identified cases which we considered to overlap, the next question is to determine the proper home for those cases.
To do this, each member of the research team labelled, and then we assigned labels based on the majority vote.
Thus, we have some semi-rigorous means to claim that when a case appears to be roughly equiprobable between docTypes, we know which docType SHOULD have that concept.}

Having identified cases which we considered to overlap, the next question is to determine the proper home for those cases.
To do this, each author labelled each case as either privacy-related or not.
They performed this binary classification based on reading the definitions or one-paragraph explanations given by the ToSDR team on their website.
Due to explicit reference to privacy-related issues, certain cases, such as \emph{``Some personal data may be kept for business interests or legal obligations,''} lend themselves to clear and unambiguous labelling.
However, given the inherent subjectivity of individual opinions, the researchers' labels differed at times, such as for cases with less specific information (e.g., \emph{``Pseudonyms are allowed''} or \emph{``Only necessary logs are kept by the service to ensure quality''}.

We annotated a total of 245 cases (the 246th is ``abstain"), with each case being binary-classified as either not relevant to privacy (0) or connected to privacy (1).
This resulted in a total of 980 annotations because there were 4 annotators for each example.
Then, we used Cohen's kappa to measure the consistency of the annotations and the level of agreement between the annotators, obtaining a score of 0.5481, which indicates a moderate level of agreement between the annotators.

\boldify{So now we have a bunch of labels.
How do we combine them? Kappa}

Having created varied individual labellings, we needed to combine them into a final verdict in order to determine the best acceptable category for each case.
To do this, the team employed the kappa score~\cite{cohen1960coefficient} as a statistical measure to gauge the inter-rater reliability for this binary classification task.
The kappa score serves as a robust measure of the degree of agreement between the researchers, taking into account the possibility of chance agreement.
Utilizing this score, the team was able to evaluate and consolidate the independently assigned labels, thereby arriving at a consensus decision for each case.
\section{Empirical Results
\draftStatus{D0.0}}

\subsection{RQ1 - Case Prediction
\draftStatus{D1.0}}
\label{sec:resultsCase}

\begin{table*}[htb]
    \centering
    \begin{tabular}{@{}lc|cc|cc@{}}
     \textbf{Model} & \textbf{Sampling} &
     \textbf{Accuracy} &\textbf{F1-score} &
     \textbf{Precision} & \textbf{Recall}
     \\\hline
     \roberta{} & Normal &
     \textbf{0.7547} & \textbf{0.7367}	& 0.7547 & 0.7352
     \\\hline

     \privbert{} & Normal &
     0.7550 & 0.7348 & 0.7399	& \textbf{0.7550} 
     \\\hline
     
     \privbert{} &  Oversample &
     0.7378	& 0.7315 & \textbf{0.7568} & 0.7378
     \\\hline
     
     \roberta{} & Oversample &
     0.7119 & 0.7088 & 0.7485 & 0.7119 
     \\\hline
     
    \rf{} & Oversample &
    0.6504 & 0.6311	& 0.6584 & 0.6504
    \\\hline

    \rf{} & Normal &
    0.6367 & 0.5932 & 0.6027 & 0.6367
    \\\hline

    \svm{} & Normal &
    0.4600 & 0.4600 & 0.4500 & 0.4600
    \\\hline
    
    \svm{} & Oversample &
    0.4428 & 0.4397 & 0.4573 & 0.4428
\\\hline
    GPTv4Turbo{} & Normal &
    0.5831 & 0.5781 & 0.6747 & 0.5831
    \\\hline
    
    \end{tabular}
    \caption{Macro Averaging Evaluation Metrics for Every Model for the Case Classification task, ordered by F1-score .}
    \label{table:caseF1}
\end{table*}

\boldify{This section stands on Table~\ref{table:caseF1} mostly (with a little extra info about compute time). There are 3 main points to see here: 1: \roberta{} wins}

As Table~\ref{table:caseF1} shows, \roberta{} model trained with Normal data performs the best with an F-1 score of 0.7367 for the Case classification task with 246 labels.
Meanwhile, \privbert{} is on par with \roberta{} giving an F1-score of 0.7348. 

\boldify{2: Transformers consistently beat the classic approaches}

Table~\ref{table:caseF1} also shows that \rf{} and \svm{} models are substantially and consistently outclassed by the transformer models \roberta{} and \privbert{}, highlighting the effectiveness of deep learning for complex textual classification tasks.
Among the traditional machine learning models, \rf{} outperforms \svm{}.

\boldify{3: but they have an increased computational cost}

But, the performance improvement from transformer models such as \roberta{} and \privbert{} comes at a cost: a considerable increase in compute time compared to traditional machine learning algorithms \svm{} and \rf{}.
All our training tasks ran on NVIDIA RTX A4500 GPUs.
On this hardware, training with our hyperparameters in the Case classification task with 245 labels, \roberta{} takes 65 minutes on `Normal' data and 210 minutes on `Oversampled'.
\privbert{} trains a little faster, both on `Normal' (60 minutes), and `Oversampled' data (180 minutes).
Meanwhile, on the simpler DocType classification task with 5 labels, the \roberta{} and \privbert{} models behave more similarly, requiring 19 minutes on `Normal' data and 45 minutes on `Oversampled' data.

\subsection{RQ2 - Overlap Quantification
\draftStatus{D1.0}}
\label{secOverlap}

\begin{table*}[htb]
    \centering
    \begin{tabular}{@{}lc|cc|cc@{}}
     \textbf{Model} & \textbf{Sampling} &
     \textbf{Accuracy} &\textbf{F1-score} &
     \textbf{Precision} & \textbf{Recall}\\
     \hline
     \roberta{} & Normal&
     0.8120 & \textbf{0.7977} & \textbf{0.7916}	& 0.8109
     \\\hline

     \privbert{} &  Oversample &
     \textbf{0.8129} & 0.7934 & 0.7853 & \textbf{0.8129} 
     \\\hline
     
     \privbert{} & Normal&
     0.8068 & 0.7911	& 0.7878 & 0.8068
     \\\hline
     
     \roberta{} & Oversample &
     0.7869 & 0.7751	& 0.7686 & 0.7869
     \\\hline
     
    \svm{} & Normal&
    0.7880 & 0.7593	& 0.7428 & 0.7880
    \\\hline

    \rf{} & Oversample &
    0.7866 & 0.7578	& 0.7512 & 0.7866
    \\\hline
    
    \svm{} & Oversample &
    0.7259 & 0.7333 & 0.7467 & 0.7259
    \\\hline
    
    \rf{} & Normal&
    0.6509 & 0.6108	& 0.6163 & 0.6509
    
    \\\hline
    GPTv4Turbo{} & Normal &
    0.8596 & 0.8483 & 0.8784 & 0.8596
    \\\hline
    
    \end{tabular}
    \caption{Macro Averaging Evaluation Metrics for Every Model on the DocType Classification task, ordered by F1-score.}
    \label{table:policyF1}
\end{table*}

\boldify{Here, we look to Table~\ref{table:policyF1} and 1: see that concept overlap seems to be present, but is not severe}

As Table~\ref{table:policyF1} illustrates, our best F1 and accuracy scores are in the .78--.81 range for the DocType classification task.
Thus, it seems reasonable to conclude that there is \emph{some} concept overlap among policy documents, but it is not \emph{severe}.
On one hand, if concept overlap was nonexistent, these numbers would be closer to one.
On the other, if concept overlap was complete, these numbers would be closer to .2, reflecting random guessing with 5 classes.

\boldify{2: we see that all the observations made in the previous section are also present again}

Table~\ref{table:policyF1} also shows all the trends we observed in Section~\ref{sec:resultsCase}.
\roberta{} is best, with \privbert{} a close second, both take more computing and perform better than the classical models.

\boldify{3: ...with the caveat that classic approaches are more competitive (probably due to the simpler labeling space)}

However, on this task, we must offer a caveat: the classical models are \emph{much} more competitive with the transformers.
We believe this is due to the simpler labeling space of 5 classes instead of 246 as in the Case classification task.

\boldify{4: Oversampling appears not to have worked, let's take a deeper dive with Table~\ref{tab:confusion}.
GenderShades can offer some insight here}

Consulting Tables~\ref{table:caseF1} and \ref{table:policyF1}, it appears that Oversampling did not help model performance.
While \rf{} is a notable exception, and \privbert{} on one task, the models with Normal sampling perform better.
Table~\ref{tab:confusion} takes a deeper dive into how oversampling affected the models, in particular highlighting how the performance went to zero for the `Data Policy' doctype when Oversampling.
In general, we would expect Oversampling to do a little better on minority classes (and maybe a little worse on majority classes).
However, our oversampling is clearly not sufficient, suggesting the need for a better sampling approach, data augmentation, or additional data collection, as \briefCite{buolamwini18gendershades} recommended for a similar data distribution problem in a different domain.

Initially, we were hoping to provide the output of a statistical test comparing the distributions of case labels between the two document types.
However, we were not able to find one that was capable of handling the categorical data we have.
Thus, we have created and computed the following loss function for binary classification (here the classes are \termsofserviceLong{} and Privacy Policy documents):
\begin{equation}
Loss = \frac{1}2 \sum_{c} | f_c(PP) - f_c(ToS) | 
\label{eqLossFn}
\end{equation}
...where $c$ is a case, and the input is normalized such that each class has a function $f_c(\cdot)$ returning ``fraction of data receiving that label'' (note that this is NOT a count, but a fraction).

Equation~\ref{eqLossFn} results in a quantity varying from 0 to 1, where 0 indicates the distributions are the same, and 1 indicates the distributions are non-overlapping.
Had we not normalized the samples, the high end of this quantity would be unbounded.
The loss computation from our best-performing model (RoBERTa) yields 0.6146, indicating that the distributions are a bit on the dissimilar side.
\subsection{RQ3 - Cases that Overlap
\draftStatus{D0.3}}

\begin{table*}[htb]
    \centering
    \begin{tabular}{@{}l|ccccc||ccccc@{}}
    & \termsofservice{}	& PP	& CP	& DP	& ?P		
    & Accuracy	&Precision	&Recall	&F1-Score	& Support\\
    \hline
    \termsofservice{} &	\textbf{1300}	& 133 &	2 & 2	& 28	
     & 0.89&	0.80	&0.89	& 0.84&	1465 \\ 
    
    PP	&228	&\textbf{1581}&	39	&6	&21
    &	0.84	&0.85	&0.84	&0.85	&1875\\
    
    CP	&6	&52	&\textbf{42}	&0	&1
    &0.42	&0.50&	0.42&	0.45&	101\\
    
    DP&	3	&12	&0&	\textbf{1}	&0	&	0.06&	0.05	&0.06	&0.05&	16\\
    
    ?P&	95	&73	&1	&13	&\textbf{17}	&	0.09	&0.25	&0.09	&0.13&	199\\
    \end{tabular}

    \vspace{10pt}

    \begin{tabular}{@{}l|ccccc||ccccc@{}}
    & \termsofservice{}	& PP	& CP	& DP	& ?P		
    & Accuracy	&Precision	&Recall	&F1-Score	& Support\\
    \hline
    	
\termsofservice{}	&\textbf{1281}	&145&	3	&3&	33	&	0.87&	0.79	&0.87	&0.83	&1465\\
PP	&228	&\textbf{1586}	&51	&5	&5	&	0.85	&0.84&	0.85	&0.85&	1875\\
CP&	3	&56	&\textbf{40}	&2	&0	&	0.40	&0.42&	0.4	&0.41	&101\\
DP&	3&	56	&40&	\textbf{2}	&0	&	0.02&	0&	0&	0	&16\\
?P	&100&	78	&1	&10	&\textbf{10}		&0.05	&0.21&	0.05&	0.08	&199\\
    \end{tabular}
    \caption{(\textbf{top:}) Macro Averaging Confusion Matrix and classification report for the \roberta{} DocType classifier with Normal sampling.
    (\textbf{bottom:}) All the same, but with Oversampling
    \\ \textbf{PP} = Privacy Policy, \textbf{CP} = Cookie Policy, \textbf{DP} = Data Policy, \textbf{?P} = Unknown Policy}
    \label{tab:confusion}
\end{table*}

\begin{figure*}
    \centering
    \includegraphics[width=\linewidth]{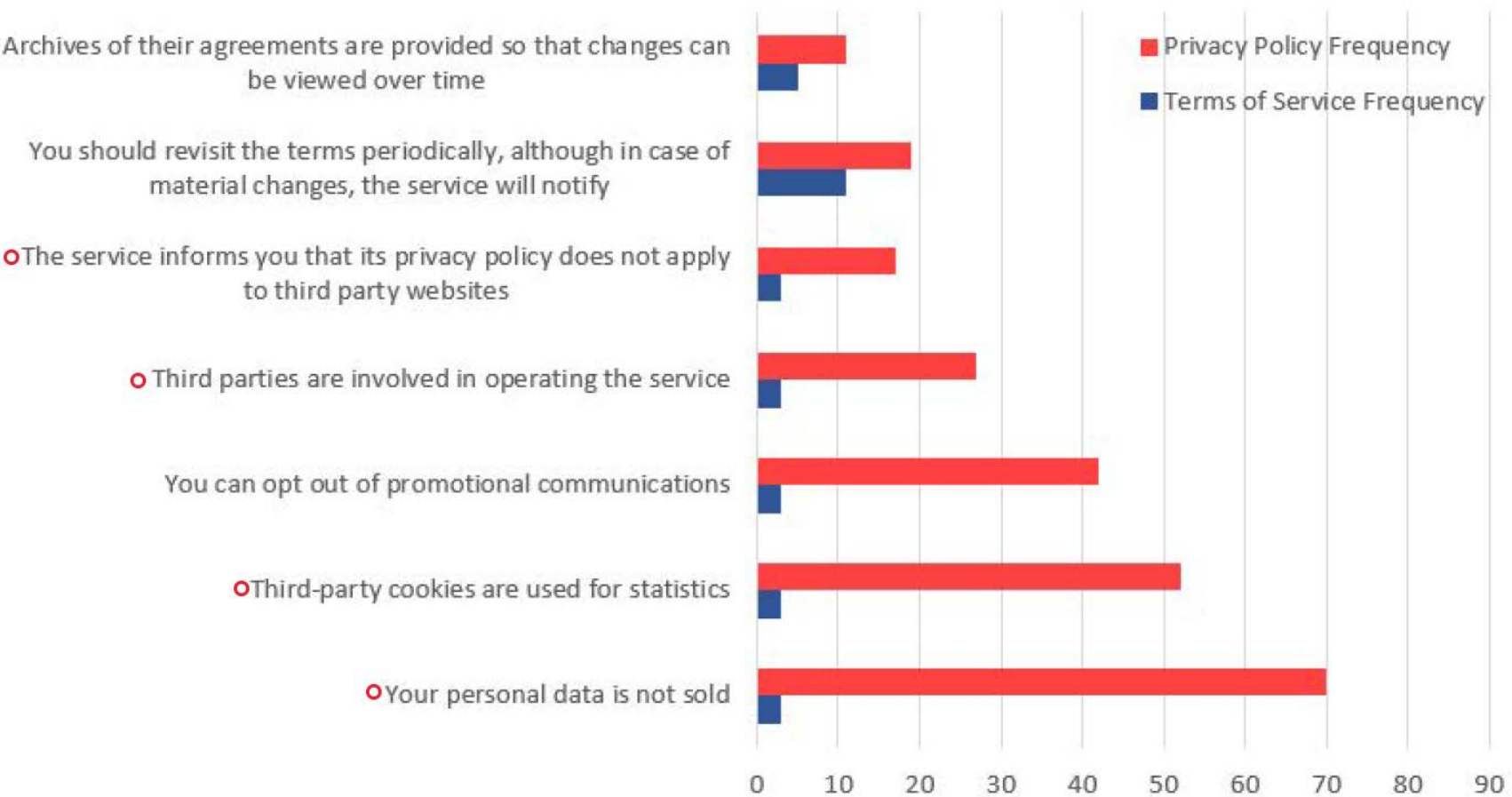}
    
    \caption{Overlapping cases where most instances appear in a Privacy Policy.
    Cases that we labelled as ``Privacy Related'' are marked with a red $\circ$ (4 out of 7 cases).}
    \label{fig:encroachmentPP}

\end{figure*}

\begin{figure*}
    \centering
    \includegraphics[width=\linewidth]{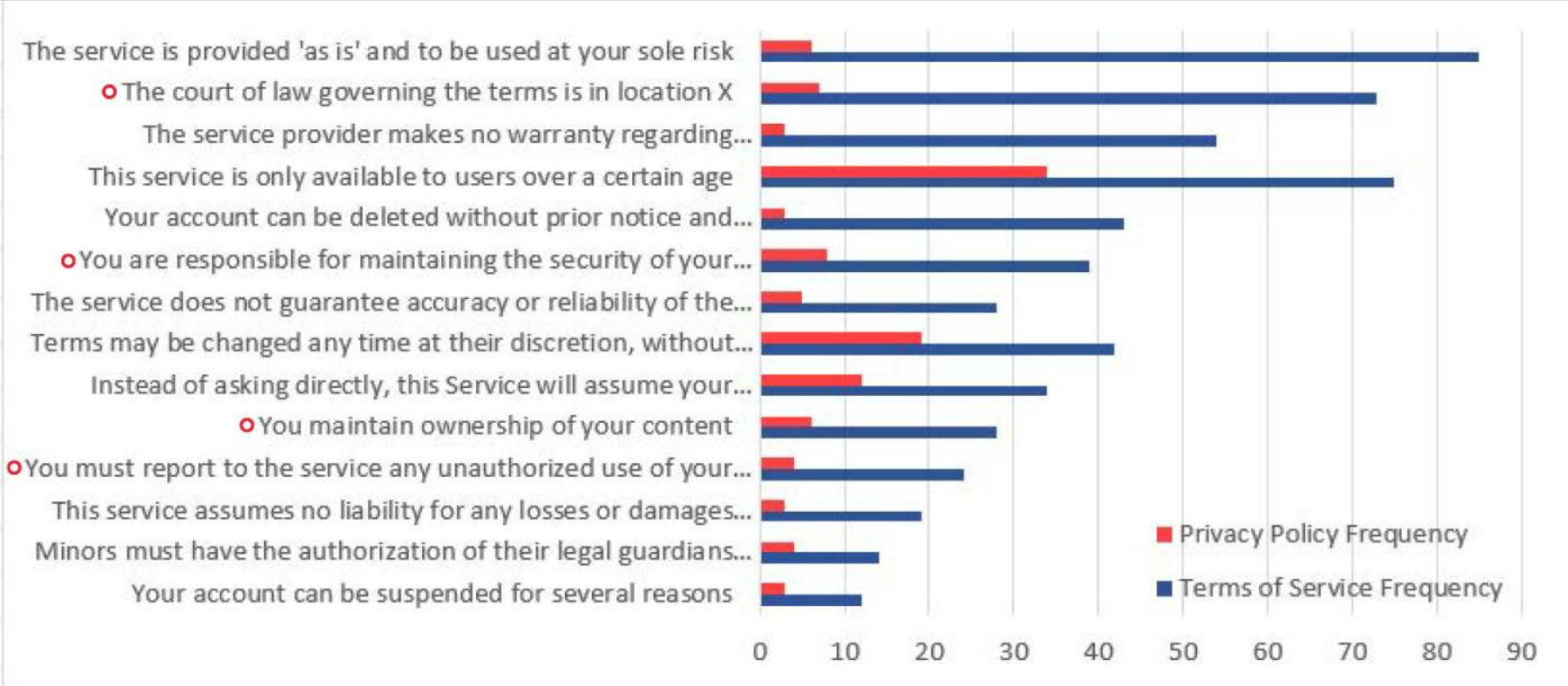}
    
    \caption{Overlapping cases where most instances appear in a Terms of Service.
    Cases that we labelled as ``Privacy Related'' are marked with a red $\circ$ (4 out of 14 cases).}
    \label{fig:encroachmentToS}
\end{figure*}

\begin{figure*}
    \includegraphics[width=\linewidth]{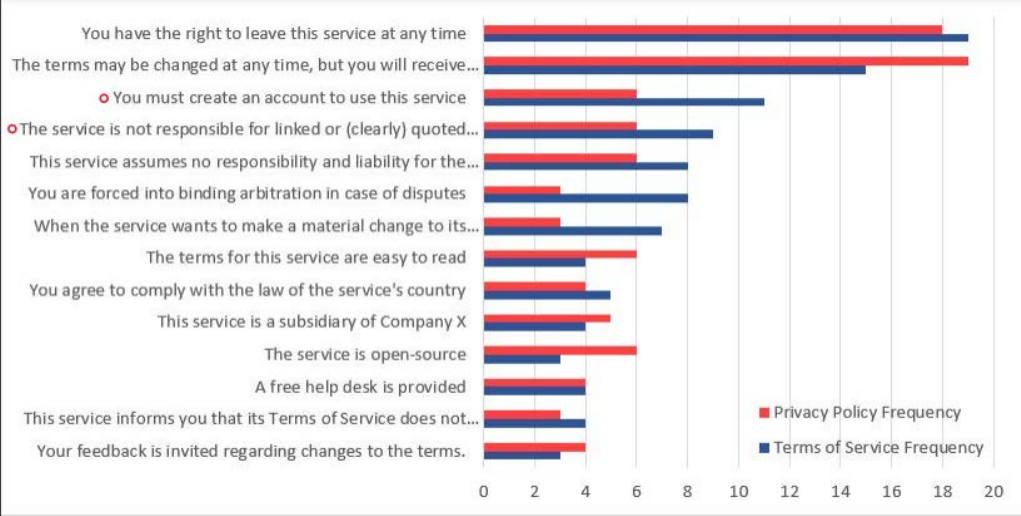}
    
    \caption{Frequency of overlapping cases between Privacy Policy and Terms of Service.
    Cases that we labelled as ``Privacy Related'' are marked with a red $\circ$ (2 out of 14 cases).}
    \label{fig:freq_case_policy}
\end{figure*}

\boldify{Here we switch focus over to Figures \ref{fig:encroachmentPP}--\ref{fig:freq_case_policy}, and notice the following: 
1: After filtering (3+ instances on each side, no "has a date"), we can split the overlapping cases into 3x regimes: PP dominant, \termsofservice{} dominant, and contested (each have a figure.
Our splitpoints were $\pm 5$ on the difference between the two frequencies.)}

For this RQ, we switch focus over to Figures~\ref{fig:encroachmentPP}--\ref{fig:freq_case_policy}, which depict all\footnote{%
All except one case (\emph{``There is a date of the last update of the agreements''}), which had about 150 occurrences in both docTypes, dominating chart axes.}
of the cases that have three or more examples appearing in the test data for both privacy policies and \termsofservice{}.
After establishing this set, we sorted it by the difference between appearances in each docType, and split the data into 3 regimes based on this difference being $>5$ (Figure~\ref{fig:encroachmentPP}), $<-5$ (Figure~\ref{fig:encroachmentToS}), or $<5$ \&\& $>-5$ (Figure~\ref{fig:freq_case_policy}).
We have marked each case that our human-labelling described in Section~\ref{sec:overlap} found to be ``privacy-related'' with a red circle on the left of the description.

\boldify{2: Starting with PP dominant (fig4), we see N concepts which usually appear in PP docs.
Of these, most were things our human-labelling effort tagged as privacy related ($|X|/|Y|$, X=...)}

Figure~\ref{fig:encroachmentPP} lists concepts tending to mostly appear in privacy policies.
Note that most of these cases (4/7) are marked with the red circle indicating that we found them to be ``privacy-related,'' as one might expect of this document type.
The remaining cases, which seem potentially misplaced are: 
\emph{``Archives of their agreements are provided so that changes can be viewed over time''};
\emph{``You should revisit the terms periodically, although in case of material changes, the service will notify''}; and
\emph{``You can opt out of promotional communications''}.

\boldify{3: Moving to the \termsofservice{} dominant (fig5), we see M concepts which usually appear in \termsofservice{} docs.
Of these, most were NOT tagged as privacy-related ($|X|/|Y|$, X=...)}

Moving on to Figure~\ref{fig:encroachmentToS}, we see concepts that tended to appear in \termsofservice{}.
As expected, fewer of these cases are marked with the red circle indicating that we found them to be ``privacy-related'' (4/14).
The remaining cases, which seem potentially misplaced, are: 
\emph{``The court of law governing the terms is in location X''};
\emph{``You are responsible for maintaining the security of your account and for the activities on your account''};
\emph{``You maintain ownership of your content''}; and
\emph{``You must report to the service any unauthorized use of your account or any breach of security''}.
This indicates that the privacy policies are encroached with non-privacy-related content.
\boldify{4: moving into the "contested" territory, we see the following additional concepts which the community seems to be a bit unclear if they fall into one bucket or the other.}

Last, Figure~\ref{fig:freq_case_policy} shows the most interesting group: concepts that appeared in both document types in roughly equal measure.
These are the concepts one might consider ``contested,'' in that the community of policy writers seems to disagree about which document should exclusively feature this information.
Key privacy-related concepts in this group include \emph{``You must create an account to use this service''} and \emph{``The service is not responsible for linked or (clearly) quoted content from third-party content providers''}.

\boldify{5: In sum, this is the general flavor of cases that appear to be confusing to the document WRITERS, as measured by our language models.}

In sum, the cases listed in this section are subject to some disagreement (due to error or conscious objection) on where to locate certain information.
In particular, privacy policies seems to be encroaching on content better suited for \termsofservice{} because 3/7 concepts are ``out of place'' in PP-dominated data (Figure~\ref{fig:encroachmentPP}), 4/14 out of place in ToS-dominated data (Figure~\ref{fig:encroachmentToS}), and 12 contested concepts are out of place in PP (Figure~\ref{fig:freq_case_policy})
By raising awareness of this disagreement, we hope the community of policy writers can be clearer for the humans that read their documents, as well as the NLP models that assist the humans.

\section{Discussion
\draftStatus{D1.0}}
\label{secDiscussion}

\subsection{To what extent are our models ``good enough'' to build a tool that internet users might find useful?}

\boldify{We think so, but user studies are the real answer.
Next, I'll provide a little more detail to show that I have thought about it more than "do user studies"}

While we think our models are good enough to build a useful tool, user studies will be the ultimate arbiter of success on this question.
Once our tool reaches a greater level of maturity, we envision conducting qualitative user studies examining previously existing workflows for understanding policy documents, as well as workflows arising during the use of our system.
Next, quantitative user studies would measure the effectiveness of our tool along dimensions such as ``knowledge about the contents of the documents'' (evaluated pre-then-post) and ``time to attain that knowledge,'' as found in prior work~\cite{golbeck2016}.
There are several approaches that are likely to improve upon our results, such as paying for more data labeling or performing data augmentation (in rare cases, in particular).

\subsection{Why use a simplification strategy, as opposed to summarization?}

\boldify{Reasons for this choice are mostly to make our later goal of scoring feasible and also that human knowledge in the form of taxonomy and labeling is very useful.
Last, our results might differ @somehow@ had we chosen to use summarization.}

Summarization likely has a place in helping humans understand policy documents, but it is not clear how one might numerically ``score'' the output summary concerning the terms' favorability for the user.
Since our goal is to ultimately provide such scores, we chose to adopt a classification problem framing that makes numeric scoring straightforward by assigning a scalar value to each case.
Further, by using the ToSDR taxonomy and labeled data, we leverage a lot of human knowledge that summarization would not effectively employ.
In sum, had we used summarization, our results would not have the necessary granularity for scoring.

\subsection{To what extent are overlap analysis results driven by document similarity, as opposed to the quality of the model used?}

\boldify{We basically did 3 things here: 1) used multiple sources of data for triangulation, 2) evaluated several models, and 3) followed an established pattern}

We approached the overlap in two ways (by case and by docType) to achieve greater triangulation for those results.
Ideally, one might measure document similarity by looking at the best achievable accuracy by classifying the source document type \emph{over the space of all possible models}.
However, given time and computing limitations, we restricted ourselves to the models presented in this paper.
Last, the type of argumentation found in this paper has some precedent, for example, work by Pozen et al.~\cite{pozen2019} that measures partisanship via classifiers' accuracy at predicting the part affiliation of speakers in the congressional record.

We opted to start with a very naïve off-the-shelf technique for our sampling strategy.
While choosing an over/under/hybrid sampling strategy based on a given problem is very context-dependent, our results would likely be improved by using class elimination, more data collection, data augmentation, or a more complex sampling strategy (like SMOTE~\cite{chawla2002smote} or its variants).
Ultimately, we favored using the off-the-shelf ToSDR taxonomy for simplicity and replication.
That said, the labels contain negations (e.g., service will sell your data vs will not sell), or other forms of parameterized labels (e.g., the applicable law is GDPR/CCPA/...).
Such parameterized labeling ensures holistically considering all actions in a certain segment~\cite{qiu2023calpric}. Treating the task as a classification on ToSDR taxonomy certainly does have the cost of discarding semantic information about the relatedness of such cases.

We limited the scope of this paper to a small set of four models, which share two main properties: first, some expectation that the model might perform reasonably on the task; second, sufficient access to the model to allow us to offer users explainability or interpretability when the model becomes part of an end-to-end pipeline with a user interface.
In particular, if the model says \textit{``this terms of service is good, sign it.''}, a user might want more information, such as pointers to parts of the document that the classifier identified.
This means that the BERT family becomes our state-of-the-art, because LLMs are precluded due to insufficient access for planned future work, though they may perform some of the individual tasks presented in this paper much better.
Additionally, Random Forest and SVM are common baselines that also support techniques from eXplainable AI (XAI).

With huge number of labels and a long tail distribution, there is no perfect metric to evaluate models.
The model will overfit given the smaller number of data points for each case label.
One of the reasons to prefer f1 over accuracy is the class imbalance in data.

We have modelled this task as a single-instance classification problem.
However, the presumption that each sentence belongs to only one of the 246 classes will not always hold true in the world of document classification, especially when dealing with complex legal documents.
To address this, one could segment at a more granular level (e.g., phrase) or consider the data as multi-instance.
Unfortunately, that is not the way already labelled data exist, and so we stuck with the single-instance framing.

\section{Conclusion
\draftStatus{D1.0}}

As privacy concerns become increasingly central to people’s everyday lives, it is important to understand how natural language processing models can be used to make their lives better.
With the central objective of this paper being to break down the legalese in a manner that is concise and understandable to everyone, we hope that this automated summarization of policy documents will help to simplify the verbosity of the original documents.

From a service provider's perspective, we highlighted the redundancies and the overlaps in the different documents that need to be unique to certain texts.
By training the model on the dataset of over 10,000 annotations, the RoBERTa model gives back an F1-score of 0.74.
The transformer-based models invariably perform better than their traditional counterparts with an increased efficacy coupled with higher computational time.
Results also suggest some commonalities in the policy documents suggesting a significant indulgent overlap.
These significant overlaps in the number of cases are indicative of the lack of clarity of terminologies between the two types of policy documents and their contents.
This study identifies those overlaps and directs the policymaker to make appropriate changes in future versions of the document.

\bibliography{0-main}


\begin{thebibliography}{44}


\ifx \showCODEN    \undefined \def \showCODEN     #1{\unskip}     \fi
\ifx \showDOI      \undefined \def \showDOI       #1{#1}\fi
\ifx \showISBNx    \undefined \def \showISBNx     #1{\unskip}     \fi
\ifx \showISBNxiii \undefined \def \showISBNxiii  #1{\unskip}     \fi
\ifx \showISSN     \undefined \def \showISSN      #1{\unskip}     \fi
\ifx \showLCCN     \undefined \def \showLCCN      #1{\unskip}     \fi
\ifx \shownote     \undefined \def \shownote      #1{#1}          \fi
\ifx \showarticletitle \undefined \def \showarticletitle #1{#1}   \fi
\ifx \showURL      \undefined \def \showURL       {\relax}        \fi
\providecommand\bibfield[2]{#2}
\providecommand\bibinfo[2]{#2}
\providecommand\natexlab[1]{#1}
\providecommand\showeprint[2][]{arXiv:#2}

\bibitem[Adhikari(2020)]%
        {adhikari2020automated}
\bibfield{author}{\bibinfo{person}{Andrick~Dev Adhikari}.}
  \bibinfo{year}{2020}\natexlab{}.
\newblock \emph{\bibinfo{title}{Automated Change Detection in Privacy
  Policies}}.
\newblock \bibinfo{thesistype}{Ph.\,D. Dissertation}.
  \bibinfo{school}{University of Denver}.
\newblock


\bibitem[Ahmad et~al\mbox{.}(2021)]%
        {ahmad2021intent}
\bibfield{author}{\bibinfo{person}{Wasi~Uddin Ahmad}, \bibinfo{person}{Jianfeng
  Chi}, \bibinfo{person}{Tu Le}, \bibinfo{person}{Thomas Norton},
  \bibinfo{person}{Yuan Tian}, {and} \bibinfo{person}{Kai-Wei Chang}.}
  \bibinfo{year}{2021}\natexlab{}.
\newblock \showarticletitle{Intent classification and slot filling for privacy
  policies}.
\newblock \bibinfo{journal}{\emph{arXiv preprint arXiv:2101.00123}}
  (\bibinfo{year}{2021}).
\newblock


\bibitem[Ahmad et~al\mbox{.}(2020)]%
        {ahmad2020policyqa}
\bibfield{author}{\bibinfo{person}{Wasi~Uddin Ahmad}, \bibinfo{person}{Jianfeng
  Chi}, \bibinfo{person}{Yuan Tian}, {and} \bibinfo{person}{Kai-Wei Chang}.}
  \bibinfo{year}{2020}\natexlab{}.
\newblock \showarticletitle{PolicyQA: A reading comprehension dataset for
  privacy policies}.
\newblock \bibinfo{journal}{\emph{arXiv preprint arXiv:2010.02557}}
  (\bibinfo{year}{2020}).
\newblock


\bibitem[Alabduljabbar and Mohaisen(2022)]%
        {alabduljabbar2022measuring}
\bibfield{author}{\bibinfo{person}{Abdulrahman Alabduljabbar} {and}
  \bibinfo{person}{David Mohaisen}.} \bibinfo{year}{2022}\natexlab{}.
\newblock \showarticletitle{Measuring the privacy dimension of free content
  websites through automated privacy policy analysis and annotation}. In
  \bibinfo{booktitle}{\emph{Companion Proceedings of the Web Conference 2022}}.
  \bibinfo{pages}{860--867}.
\newblock


\bibitem[Amos et~al\mbox{.}(2021)]%
        {amos2021privacy}
\bibfield{author}{\bibinfo{person}{Ryan Amos}, \bibinfo{person}{Gunes Acar},
  \bibinfo{person}{Eli Lucherini}, \bibinfo{person}{Mihir Kshirsagar},
  \bibinfo{person}{Arvind Narayanan}, {and} \bibinfo{person}{Jonathan Mayer}.}
  \bibinfo{year}{2021}\natexlab{}.
\newblock \showarticletitle{Privacy policies over time: Curation and analysis
  of a million-document dataset}. In \bibinfo{booktitle}{\emph{Proceedings of
  the Web Conference 2021}}. \bibinfo{pages}{2165--2176}.
\newblock


\bibitem[Auxier et~al\mbox{.}(2019)]%
        {auxier2019americans}
\bibfield{author}{\bibinfo{person}{Brooke Auxier},
  \bibinfo{person}{Monica~Anderson Lee~Raine}, \bibinfo{person}{Andrew Perrin},
  \bibinfo{person}{Madhu Kumar}, {and} \bibinfo{person}{Erica Turner}.}
  \bibinfo{year}{2019}\natexlab{}.
\newblock \showarticletitle{Americans’ attitudes and experiences with privacy
  policies and laws}.
\newblock \bibinfo{journal}{\emph{Pew Research Center: Internet, Science \&
  Tech}} (\bibinfo{year}{2019}).
\newblock
\urldef\tempurl%
\url{https://www.pewresearch.org/internet/2019/11/15/americans-attitudes-and-experiences-with-privacy-policies-and-laws/}
\showURL{%
\tempurl}


\bibitem[Bakos et~al\mbox{.}(2014)]%
        {bakos2014does}
\bibfield{author}{\bibinfo{person}{Yannis Bakos}, \bibinfo{person}{Florencia
  Marotta-Wurgler}, {and} \bibinfo{person}{David~R Trossen}.}
  \bibinfo{year}{2014}\natexlab{}.
\newblock \showarticletitle{Does anyone read the fine print? Consumer attention
  to standard-form contracts}.
\newblock \bibinfo{journal}{\emph{The Journal of Legal Studies}}
  \bibinfo{volume}{43}, \bibinfo{number}{1} (\bibinfo{year}{2014}),
  \bibinfo{pages}{1--35}.
\newblock


\bibitem[Bolton et~al\mbox{.}(2023)]%
        {bolton2023imbalance}
\bibfield{author}{\bibinfo{person}{Tom Bolton}, \bibinfo{person}{Tooska
  Dargahi}, \bibinfo{person}{Sana Belguith}, {and} \bibinfo{person}{Carsten
  Maple}.} \bibinfo{year}{2023}\natexlab{}.
\newblock \showarticletitle{PrivExtractor: Toward Redressing the Imbalance of
  Understanding between Virtual Assistant Users and Vendors}.
\newblock \bibinfo{journal}{\emph{ACM Trans. Priv. Secur.}}
  \bibinfo{volume}{26}, \bibinfo{number}{3}, Article \bibinfo{articleno}{31}
  (\bibinfo{date}{may} \bibinfo{year}{2023}), \bibinfo{numpages}{29}~pages.
\newblock
\showISSN{2471-2566}
\urldef\tempurl%
\url{https://doi.org/10.1145/3588770}
\showDOI{\tempurl}


\bibitem[Breiman(2001)]%
        {breiman2001random}
\bibfield{author}{\bibinfo{person}{Leo Breiman}.}
  \bibinfo{year}{2001}\natexlab{}.
\newblock \showarticletitle{Random forests}.
\newblock \bibinfo{journal}{\emph{Machine learning}}  \bibinfo{volume}{45}
  (\bibinfo{year}{2001}), \bibinfo{pages}{5--32}.
\newblock


\bibitem[Bui et~al\mbox{.}(2021)]%
        {bui2021automated}
\bibfield{author}{\bibinfo{person}{Duc Bui}, \bibinfo{person}{Kang~G Shin},
  \bibinfo{person}{Jong-Min Choi}, {and} \bibinfo{person}{Junbum Shin}.}
  \bibinfo{year}{2021}\natexlab{}.
\newblock \showarticletitle{Automated extraction and presentation of data
  practices in privacy policies}.
\newblock \bibinfo{journal}{\emph{Proceedings on Privacy Enhancing
  Technologies}} (\bibinfo{year}{2021}).
\newblock


\bibitem[Buolamwini and Gebru(2018)]%
        {buolamwini18gendershades}
\bibfield{author}{\bibinfo{person}{Joy Buolamwini} {and}
  \bibinfo{person}{Timnit Gebru}.} \bibinfo{year}{2018}\natexlab{}.
\newblock \showarticletitle{Gender Shades: Intersectional Accuracy Disparities
  in Commercial Gender Classification}. In
  \bibinfo{booktitle}{\emph{Proceedings of the 1st Conference on Fairness,
  Accountability and Transparency}} \emph{(\bibinfo{series}{Proceedings of
  Machine Learning Research}, Vol.~\bibinfo{volume}{81})},
  \bibfield{editor}{\bibinfo{person}{Sorelle~A. Friedler} {and}
  \bibinfo{person}{Christo Wilson}} (Eds.). \bibinfo{publisher}{PMLR},
  \bibinfo{pages}{77--91}.
\newblock
\urldef\tempurl%
\url{https://proceedings.mlr.press/v81/buolamwini18a.html}
\showURL{%
\tempurl}


\bibitem[Chawla et~al\mbox{.}(2002)]%
        {chawla2002smote}
\bibfield{author}{\bibinfo{person}{Nitesh~V Chawla}, \bibinfo{person}{Kevin~W
  Bowyer}, \bibinfo{person}{Lawrence~O Hall}, {and} \bibinfo{person}{W~Philip
  Kegelmeyer}.} \bibinfo{year}{2002}\natexlab{}.
\newblock \showarticletitle{SMOTE: synthetic minority over-sampling technique}.
\newblock \bibinfo{journal}{\emph{Journal of artificial intelligence research}}
   \bibinfo{volume}{16} (\bibinfo{year}{2002}), \bibinfo{pages}{321--357}.
\newblock


\bibitem[Cohen(1960)]%
        {cohen1960coefficient}
\bibfield{author}{\bibinfo{person}{Jacob Cohen}.}
  \bibinfo{year}{1960}\natexlab{}.
\newblock \showarticletitle{A coefficient of agreement for nominal scales}.
\newblock \bibinfo{journal}{\emph{Educational and psychological measurement}}
  \bibinfo{volume}{20}, \bibinfo{number}{1} (\bibinfo{year}{1960}),
  \bibinfo{pages}{37--46}.
\newblock


\bibitem[Cortes and Vapnik(1995)]%
        {cortes1995support}
\bibfield{author}{\bibinfo{person}{Corinna Cortes} {and}
  \bibinfo{person}{Vladimir Vapnik}.} \bibinfo{year}{1995}\natexlab{}.
\newblock \showarticletitle{Support-vector networks}.
\newblock \bibinfo{journal}{\emph{Machine learning}}  \bibinfo{volume}{20}
  (\bibinfo{year}{1995}), \bibinfo{pages}{273--297}.
\newblock


\bibitem[Court of Appeals, 9th Circuit(2012)]%
        {2012davis}
Court of Appeals, 9th Circuit \bibinfo{year}{2012}\natexlab{}.
\newblock \bibinfo{title}{Davis v. HSBC Bank Nevada, NA}.
\newblock , \bibinfo{numpages}{1152}~pages.
\newblock


\bibitem[Gao et~al\mbox{.}(2011)]%
        {gao2011mining}
\bibfield{author}{\bibinfo{person}{Xibin Gao}, \bibinfo{person}{Munindar~P
  Singh}, {and} \bibinfo{person}{Pankaj Mehra}.}
  \bibinfo{year}{2011}\natexlab{}.
\newblock \showarticletitle{Mining business contracts for service exceptions}.
\newblock \bibinfo{journal}{\emph{IEEE Transactions on Services Computing}}
  \bibinfo{volume}{5}, \bibinfo{number}{3} (\bibinfo{year}{2011}),
  \bibinfo{pages}{333--344}.
\newblock


\bibitem[Golbeck and Mauriello(2016)]%
        {golbeck2016}
\bibfield{author}{\bibinfo{person}{Jennifer Golbeck} {and}
  \bibinfo{person}{Matthew~Louis Mauriello}.} \bibinfo{year}{2016}\natexlab{}.
\newblock \showarticletitle{User Perception of Facebook App Data Access: A
  Comparison of Methods and Privacy Concerns}.
\newblock \bibinfo{journal}{\emph{Future Internet}} \bibinfo{volume}{8},
  \bibinfo{number}{2} (\bibinfo{year}{2016}).
\newblock
\showISSN{1999-5903}
\urldef\tempurl%
\url{https://doi.org/10.3390/fi8020009}
\showDOI{\tempurl}


\bibitem[Harkous et~al\mbox{.}(2018)]%
        {harkous2018polisis}
\bibfield{author}{\bibinfo{person}{Hamza Harkous}, \bibinfo{person}{Kassem
  Fawaz}, \bibinfo{person}{R{\'e}mi Lebret}, \bibinfo{person}{Florian Schaub},
  \bibinfo{person}{Kang~G Shin}, {and} \bibinfo{person}{Karl Aberer}.}
  \bibinfo{year}{2018}\natexlab{}.
\newblock \showarticletitle{Polisis: Automated analysis and presentation of
  privacy policies using deep learning}. In \bibinfo{booktitle}{\emph{27th
  $\{$USENIX$\}$ security symposium ($\{$USENIX$\}$ security 18)}}.
  \bibinfo{pages}{531--548}.
\newblock


\bibitem[Kost and Freytag(2012)]%
        {kost2012privacy}
\bibfield{author}{\bibinfo{person}{Martin Kost} {and}
  \bibinfo{person}{Johann~Christoph Freytag}.} \bibinfo{year}{2012}\natexlab{}.
\newblock \showarticletitle{Privacy analysis using ontologies}. In
  \bibinfo{booktitle}{\emph{Proceedings of the second ACM conference on Data
  and Application Security and Privacy}}. \bibinfo{pages}{205--216}.
\newblock


\bibitem[Lippi et~al\mbox{.}(2019)]%
        {lippi2019claudette}
\bibfield{author}{\bibinfo{person}{Marco Lippi}, \bibinfo{person}{Przemys{\l}aw
  Pa{\l}ka}, \bibinfo{person}{Giuseppe Contissa}, \bibinfo{person}{Francesca
  Lagioia}, \bibinfo{person}{Hans-Wolfgang Micklitz}, \bibinfo{person}{Giovanni
  Sartor}, {and} \bibinfo{person}{Paolo Torroni}.}
  \bibinfo{year}{2019}\natexlab{}.
\newblock \showarticletitle{CLAUDETTE: an automated detector of potentially
  unfair clauses in online terms of service}.
\newblock \bibinfo{journal}{\emph{Artificial Intelligence and Law}}
  \bibinfo{volume}{27}, \bibinfo{number}{2} (\bibinfo{year}{2019}),
  \bibinfo{pages}{117--139}.
\newblock


\bibitem[Lipton and Steinhardt(2018)]%
        {lipton2018troubling}
\bibfield{author}{\bibinfo{person}{Zachary~C. Lipton} {and}
  \bibinfo{person}{Jacob Steinhardt}.} \bibinfo{year}{2018}\natexlab{}.
\newblock \bibinfo{title}{Troubling Trends in Machine Learning Scholarship}.
\newblock
\newblock
\showeprint[arxiv]{1807.03341}~[stat.ML]


\bibitem[Liu et~al\mbox{.}(2021)]%
        {liu2021have}
\bibfield{author}{\bibinfo{person}{Shuang Liu}, \bibinfo{person}{Baiyang Zhao},
  \bibinfo{person}{Renjie Guo}, \bibinfo{person}{Guozhu Meng},
  \bibinfo{person}{Fan Zhang}, {and} \bibinfo{person}{Meishan Zhang}.}
  \bibinfo{year}{2021}\natexlab{}.
\newblock \showarticletitle{Have you been properly notified? automatic
  compliance analysis of privacy policy text with GDPR article 13}. In
  \bibinfo{booktitle}{\emph{Proceedings of the Web Conference 2021}}.
  \bibinfo{pages}{2154--2164}.
\newblock


\bibitem[Liu et~al\mbox{.}(2019)]%
        {liu2019roberta}
\bibfield{author}{\bibinfo{person}{Yinhan Liu}, \bibinfo{person}{Myle Ott},
  \bibinfo{person}{Naman Goyal}, \bibinfo{person}{Jingfei Du},
  \bibinfo{person}{Mandar Joshi}, \bibinfo{person}{Danqi Chen},
  \bibinfo{person}{Omer Levy}, \bibinfo{person}{Mike Lewis},
  \bibinfo{person}{Luke Zettlemoyer}, {and} \bibinfo{person}{Veselin
  Stoyanov}.} \bibinfo{year}{2019}\natexlab{}.
\newblock \showarticletitle{Roberta: A robustly optimized bert pretraining
  approach}.
\newblock \bibinfo{journal}{\emph{arXiv preprint arXiv:1907.11692}}
  (\bibinfo{year}{2019}).
\newblock


\bibitem[Lukose et~al\mbox{.}(2022)]%
        {lukose2022privacy}
\bibfield{author}{\bibinfo{person}{Emilia Lukose}, \bibinfo{person}{Suparna
  De}, {and} \bibinfo{person}{Jon Johnson}.} \bibinfo{year}{2022}\natexlab{}.
\newblock \showarticletitle{Privacy Pitfalls of Online Service Terms and
  Conditions: a Hybrid Approach for Classification and Summarization}. In
  \bibinfo{booktitle}{\emph{Proceedings of the Natural Legal Language
  Processing Workshop 2022}}. \bibinfo{pages}{65--75}.
\newblock


\bibitem[Maronick et~al\mbox{.}(2014)]%
        {maronick2014consumers}
\bibfield{author}{\bibinfo{person}{Thomas~J Maronick} {et~al\mbox{.}}}
  \bibinfo{year}{2014}\natexlab{}.
\newblock \showarticletitle{Do consumers read terms of service agreements when
  installing software?-a two-study empirical analysis}.
\newblock \bibinfo{journal}{\emph{International Journal of Business and Social
  Research}} \bibinfo{volume}{4}, \bibinfo{number}{6} (\bibinfo{year}{2014}),
  \bibinfo{pages}{137--145}.
\newblock


\bibitem[Melicher et~al\mbox{.}(2016)]%
        {melicher2016preferences}
\bibfield{author}{\bibinfo{person}{William Melicher}, \bibinfo{person}{Mahmood
  Sharif}, \bibinfo{person}{Joshua Tan}, \bibinfo{person}{Lujo Bauer},
  \bibinfo{person}{Mihai Christodorescu}, {and} \bibinfo{person}{Pedro~Giovanni
  Leon}.} \bibinfo{year}{2016}\natexlab{}.
\newblock \showarticletitle{Preferences for Web Tracking}.
\newblock \bibinfo{journal}{\emph{Proceedings on Privacy Enhancing
  Technologies}} \bibinfo{volume}{2016}, \bibinfo{number}{2}
  (\bibinfo{year}{2016}), \bibinfo{pages}{1--20}.
\newblock


\bibitem[Obar and Oeldorf-Hirsch(2020)]%
        {obar2020biggest}
\bibfield{author}{\bibinfo{person}{Jonathan~A Obar} {and} \bibinfo{person}{Anne
  Oeldorf-Hirsch}.} \bibinfo{year}{2020}\natexlab{}.
\newblock \showarticletitle{The biggest lie on the internet: Ignoring the
  privacy policies and terms of service policies of social networking
  services}.
\newblock \bibinfo{journal}{\emph{Information, Communication \& Society}}
  \bibinfo{volume}{23}, \bibinfo{number}{1} (\bibinfo{year}{2020}),
  \bibinfo{pages}{128--147}.
\newblock


\bibitem[Perera and Perera(2021)]%
        {perera2021barrister}
\bibfield{author}{\bibinfo{person}{Thenuka Perera} {and} \bibinfo{person}{Theja
  Perera}.} \bibinfo{year}{2021}\natexlab{}.
\newblock \showarticletitle{Barrister-Processing and Summarization of Terms \&
  Conditions/Privacy Policies}. In \bibinfo{booktitle}{\emph{2021 6th
  International Conference for Convergence in Technology (I2CT)}}. IEEE,
  \bibinfo{pages}{1--7}.
\newblock


\bibitem[Pilton et~al\mbox{.}(2021)]%
        {pilton2021evaluating}
\bibfield{author}{\bibinfo{person}{Callum Pilton}, \bibinfo{person}{Shamal
  Faily}, {and} \bibinfo{person}{Jane Henriksen-Bulmer}.}
  \bibinfo{year}{2021}\natexlab{}.
\newblock \showarticletitle{Evaluating privacy-determining user privacy
  expectations on the web}.
\newblock \bibinfo{journal}{\emph{computers \& security}}
  \bibinfo{volume}{105} (\bibinfo{year}{2021}), \bibinfo{pages}{102241}.
\newblock


\bibitem[Pozen et~al\mbox{.}(2019)]%
        {pozen2019}
\bibfield{author}{\bibinfo{person}{David~E. Pozen}, \bibinfo{person}{Eric~L.
  Talley}, {and} \bibinfo{person}{Julian Nyarko}.}
  \bibinfo{year}{2019}\natexlab{}.
\newblock \showarticletitle{A Computational Analysis of Constitutional
  Polarization}.
\newblock \bibinfo{journal}{\emph{International Journal of Business and Social
  Research}}  \bibinfo{volume}{105} (\bibinfo{year}{2019}),
  \bibinfo{pages}{1--84}.
\newblock
\urldef\tempurl%
\url{https://papers.ssrn.com/sol3/papers.cfm?abstract_id=3351339}
\showURL{%
\tempurl}


\bibitem[Qiu et~al\mbox{.}(2023)]%
        {qiu2023calpric}
\bibfield{author}{\bibinfo{person}{Wenjun Qiu}, \bibinfo{person}{David Lie},
  {and} \bibinfo{person}{Lisa Austin}.} \bibinfo{year}{2023}\natexlab{}.
\newblock \showarticletitle{Calpric: Inclusive and Fine-grain Labeling of
  Privacy Policies with Crowdsourcing and Active Learning}. In
  \bibinfo{booktitle}{\emph{32nd USENIX Security Symposium (USENIX Security
  23)}}. \bibinfo{pages}{1055--1072}.
\newblock


\bibitem[Ravichander et~al\mbox{.}(2019)]%
        {ravichander2019question}
\bibfield{author}{\bibinfo{person}{Abhilasha Ravichander},
  \bibinfo{person}{Alan~W Black}, \bibinfo{person}{Shomir Wilson},
  \bibinfo{person}{Thomas Norton}, {and} \bibinfo{person}{Norman Sadeh}.}
  \bibinfo{year}{2019}\natexlab{}.
\newblock \showarticletitle{Question answering for privacy policies: Combining
  computational and legal perspectives}.
\newblock \bibinfo{journal}{\emph{arXiv preprint arXiv:1911.00841}}
  (\bibinfo{year}{2019}).
\newblock


\bibitem[Richardson(2023)]%
        {beautifulSoup}
\bibfield{author}{\bibinfo{person}{Leonard Richardson}.}
  \bibinfo{year}{2023}\natexlab{}.
\newblock \bibinfo{title}{Beautiful Soup 4}.
\newblock
\newblock
\urldef\tempurl%
\url{https://pypi.org/project/beautifulsoup4/}
\showURL{%
\tempurl}


\bibitem[Robinson and Zhu(2020)]%
        {robinson2020beyond}
\bibfield{author}{\bibinfo{person}{Eric~P Robinson} {and}
  \bibinfo{person}{Yicheng Zhu}.} \bibinfo{year}{2020}\natexlab{}.
\newblock \showarticletitle{Beyond “I Agree”: Users’ Understanding of Web
  Site Terms of Service}.
\newblock \bibinfo{journal}{\emph{Social Media+ Society}} \bibinfo{volume}{6},
  \bibinfo{number}{1} (\bibinfo{year}{2020}),
  \bibinfo{pages}{2056305119897321}.
\newblock


\bibitem[Sadeh et~al\mbox{.}(2013)]%
        {sadeh2013usable}
\bibfield{author}{\bibinfo{person}{Norman Sadeh}, \bibinfo{person}{Alessandro
  Acquisti}, \bibinfo{person}{Travis~D Breaux}, \bibinfo{person}{Lorrie~Faith
  Cranor}, \bibinfo{person}{Aleecia~M McDonald}, \bibinfo{person}{Joel~R
  Reidenberg}, \bibinfo{person}{Noah~A Smith}, \bibinfo{person}{Fei Liu},
  \bibinfo{person}{N~Cameron Russell}, \bibinfo{person}{Florian Schaub},
  {et~al\mbox{.}}} \bibinfo{year}{2013}\natexlab{}.
\newblock \showarticletitle{The usable privacy policy project}.
\newblock In \bibinfo{booktitle}{\emph{Technical report, Technical Report,
  CMU-ISR-13-119}}. \bibinfo{publisher}{Carnegie Mellon University}.
\newblock


\bibitem[Srinath et~al\mbox{.}(2021)]%
        {srinath2021privaseer}
\bibfield{author}{\bibinfo{person}{Mukund Srinath},
  \bibinfo{person}{Soundarya~Nurani Sundareswara}, \bibinfo{person}{C~Lee
  Giles}, {and} \bibinfo{person}{Shomir Wilson}.}
  \bibinfo{year}{2021}\natexlab{}.
\newblock \showarticletitle{PrivaSeer: A Privacy Policy Search Engine}. In
  \bibinfo{booktitle}{\emph{21st International Conference on Web Engineering,
  ICWE 2021}}. Springer Science and Business Media Deutschland GmbH,
  \bibinfo{pages}{286--301}.
\newblock


\bibitem[Terms(2023)]%
        {gdprGuidelines}
\bibfield{author}{\bibinfo{person}{Privacy Terms}.}
  \bibinfo{year}{2023}\natexlab{}.
\newblock \bibinfo{title}{Privacy Policy vs Terms and Conditions}.
\newblock
\newblock
\urldef\tempurl%
\url{https://privacyterms.io/privacy/privacy-policy-vs-terms-and-conditions/}
\showURL{%
\tempurl}


\bibitem[{Terms of Service; Didn't Read}(2012)]%
        {tosdr}
\bibfield{author}{\bibinfo{person}{{Terms of Service; Didn't Read}}.}
  \bibinfo{year}{2012}\natexlab{}.
\newblock \bibinfo{title}{Project Website for `Terms of Service; Didn't Read'}.
\newblock
\newblock
\urldef\tempurl%
\url{https://tosdr.org/}
\showURL{%
\tempurl}


\bibitem[Tesfay et~al\mbox{.}(2018)]%
        {tesfay2018privacyguide}
\bibfield{author}{\bibinfo{person}{Welderufael~B Tesfay},
  \bibinfo{person}{Peter Hofmann}, \bibinfo{person}{Toru Nakamura},
  \bibinfo{person}{Shinsaku Kiyomoto}, {and} \bibinfo{person}{Jetzabel Serna}.}
  \bibinfo{year}{2018}\natexlab{}.
\newblock \showarticletitle{PrivacyGuide: towards an implementation of the EU
  GDPR on internet privacy policy evaluation}. In
  \bibinfo{booktitle}{\emph{Proceedings of the Fourth ACM International
  Workshop on Security and Privacy Analytics}}. \bibinfo{pages}{15--21}.
\newblock


\bibitem[Wagner(2023)]%
        {wagner2023ages}
\bibfield{author}{\bibinfo{person}{Isabel Wagner}.}
  \bibinfo{year}{2023}\natexlab{}.
\newblock \showarticletitle{Privacy Policies across the Ages: Content of
  Privacy Policies 1996–2021}.
\newblock \bibinfo{journal}{\emph{ACM Trans. Priv. Secur.}}
  \bibinfo{volume}{26}, \bibinfo{number}{3}, Article \bibinfo{articleno}{32}
  (\bibinfo{date}{may} \bibinfo{year}{2023}), \bibinfo{numpages}{32}~pages.
\newblock
\showISSN{2471-2566}
\urldef\tempurl%
\url{https://doi.org/10.1145/3590152}
\showDOI{\tempurl}


\bibitem[Wilson et~al\mbox{.}(2016)]%
        {wilson2016creation}
\bibfield{author}{\bibinfo{person}{Shomir Wilson}, \bibinfo{person}{Florian
  Schaub}, \bibinfo{person}{Aswarth~Abhilash Dara}, \bibinfo{person}{Frederick
  Liu}, \bibinfo{person}{Sushain Cherivirala}, \bibinfo{person}{Pedro~Giovanni
  Leon}, \bibinfo{person}{Mads~Schaarup Andersen}, \bibinfo{person}{Sebastian
  Zimmeck}, \bibinfo{person}{Kanthashree~Mysore Sathyendra},
  \bibinfo{person}{N~Cameron Russell}, {et~al\mbox{.}}}
  \bibinfo{year}{2016}\natexlab{}.
\newblock \showarticletitle{The creation and analysis of a website privacy
  policy corpus}. In \bibinfo{booktitle}{\emph{Proceedings of the 54th Annual
  Meeting of the Association for Computational Linguistics (Volume 1: Long
  Papers)}}. \bibinfo{pages}{1330--1340}.
\newblock


\bibitem[Zaeem and Barber(2021)]%
        {zaeem2021large}
\bibfield{author}{\bibinfo{person}{Razieh~Nokhbeh Zaeem} {and}
  \bibinfo{person}{K~Suzanne Barber}.} \bibinfo{year}{2021}\natexlab{}.
\newblock \showarticletitle{A Large Publicly Available Corpus of Website
  Privacy Policies Based on DMOZ.}. In \bibinfo{booktitle}{\emph{CODASPY}}.
  \bibinfo{pages}{143--148}.
\newblock


\bibitem[Zaeem et~al\mbox{.}(2018)]%
        {zaeem2018privacycheck}
\bibfield{author}{\bibinfo{person}{Razieh~Nokhbeh Zaeem},
  \bibinfo{person}{Rachel~L German}, {and} \bibinfo{person}{K~Suzanne Barber}.}
  \bibinfo{year}{2018}\natexlab{}.
\newblock \showarticletitle{Privacycheck: Automatic summarization of privacy
  policies using data mining}.
\newblock \bibinfo{journal}{\emph{ACM Transactions on Internet Technology
  (TOIT)}} \bibinfo{volume}{18}, \bibinfo{number}{4} (\bibinfo{year}{2018}),
  \bibinfo{pages}{1--18}.
\newblock


\bibitem[Zimmeck and Bellovin(2014)]%
        {zimmeck2014privee}
\bibfield{author}{\bibinfo{person}{Sebastian Zimmeck} {and}
  \bibinfo{person}{Steven~M Bellovin}.} \bibinfo{year}{2014}\natexlab{}.
\newblock \showarticletitle{Privee: An architecture for automatically analyzing
  web privacy policies}. In \bibinfo{booktitle}{\emph{23rd $\{$USENIX$\}$
  Security Symposium ($\{$USENIX$\}$ Security 14)}}. \bibinfo{pages}{1--16}.
\newblock


\end{thebibliography}
\bibliographystyle{ACM-Reference-Format}

\end{document}